\documentclass[11pt, a4paper, onecolumn, confidential, copyright, goog]{google}

\usepackage[numbers, sort&compress, round]{natbib}
\usepackage[utf8]{inputenc}
\usepackage{booktabs}
\usepackage{tabularx}
\usepackage{ragged2e}  
\usepackage{amsmath}
\usepackage{enumitem}
\usepackage{amssymb} 
\usepackage{amsthm}
\usepackage{graphicx}
\usepackage{microtype}
\usepackage{adjustbox}
\usepackage{multicol}
\usepackage{multirow}
\usepackage{colortbl} 
\usepackage{pifont}    % For checkmarks and crosses
\usepackage[most]{tcolorbox}
\usepackage{wrapfig}
\usepackage[table]{xcolor}
\usepackage[inkscapelatex=false]{svg}

\newcommand{\xmark}{\ding{55}} % Cross symbol

\usepackage[utf8]{inputenc} % allow utf-8 input
\usepackage[T1]{fontenc}    % use 8-bit T1 fonts
\usepackage{hyperref}       % hyperlinks
\usepackage{url}            % simple URL typesetting
\usepackage{amsfonts}       % blackboard math symbols
\usepackage{nicefrac}       % compact symbols for 1/2, etc.
\usepackage{xcolor}         % colors

\newcolumntype{L}{>{\raggedright\arraybackslash}X}
\newcolumntype{P}[1]{>{\raggedright\arraybackslash}p{#1}}
\usepackage[most]{tcolorbox} % Add this in the 

\uselogo{} 

\title{Forecasting Patient Trajectories from the Electronic Health Records with Action-Conditioned World Models}

\correspondingauthor{simonlee711@g.ucla.edu}

\author[1]{Simon A. Lee}
\author[1]{Jiahang Hank Sha}
\author[3]{Soomin Chung}
\author[1]{Kai Akamatsu}
\author[3]{Eran Halperin}
\author[1,2]{Jeffrey N. Chiang}

\affil[1]{UCLA Computational Medicine}
\affil[2]{UCLA Neurosurgery}
\affil[3]{New York University}

\begin{abstract}
We introduce a predictive world model for electronic health records (EHR) that represents patient histories as action-conditioned latent trajectories. Rather than modeling clinical events solely through autoregressive token prediction, our framework explicitly separates patient states from observed clinical actions and learns their evolution through a state- and action-conditioned transition operator. At its core is a state-dependent linearization predictor that propagates local changes in latent patient state, enabling time-resolved prediction, multi-step rollout, and action-conditioned simulation. Empirically, we show that generated latent trajectories preserve downstream predictive signal, temporal structure, and subgroup-level statistical relationships present in encoder-derived representations. We further decode free-running rollouts into the original clinical feature space and evaluate future clinical measurements against observed values using task-matched forecasting baselines, demonstrating that the learned dynamics retain predictive information beyond latent-space agreement. Across binary classification and time-to-event tasks, EHR-WM achieves competitive, though task-dependent, performance relative to substantially larger EHR foundation models while using a $\sim$30$\times$ smaller parameter footprint. Finally, perturbing observed action sequences produces corresponding changes in generated trajectories, demonstrating the action sensitivity of the learned dynamics. Together, these results support world modeling as a complementary framework for learning explicit, time-indexed representations of patient evolution from longitudinal EHR.
\end{abstract}

\begin{document}

\maketitle

\section{Introduction}

Predictive world models have emerged as a central abstraction in reinforcement learning and robotics, where an agent learns a compact, often latent, representation of environment dynamics \cite{ha2018world, ding2025understanding}. Formally, given a sequence of states $s_t$ and actions $a_t$, a world model parameterizes a transition operator $p_\theta(z_{t+1} \mid z_t, a_t)$, where $z_t = f_\phi(s_t)$ denotes a learned latent encoding of the observed state. This formulation enables simulation of future trajectories and supports planning through learned dynamics. Architecturally, such models combine representation learning with autoregressive temporal evolution (e.g., recurrent or transformer-based transition models). A key characteristic of this framework is the explicit factorization between controllable inputs (actions) and environment-driven evolution (states), enabling structured modeling of longitudinal systems.

A similar decomposition arises naturally in electronic health records (EHR) \cite{evans2016electronic}, where discrete clinical events can be viewed as partial observations of an underlying physiological trajectory. At each time step, the patient state consists of heterogeneous variables (e.g., vital signs, laboratory measurements, demographics, and diagnoses) that evolve over time, while treatments such as medications provide external clinical inputs that may alter subsequent evolution. This correspondence motivates casting EHR data into a $(s_t, a_t)$ framework, enabling the learning of a latent transition model $p_\theta(z_{t+1} \mid z_t, a_t)$ that represents patient evolution jointly with observed clinical actions.

In this work, we introduce a predictive world model tailored to electronic health records, formalizing patient histories as sequences of states and clinical actions in a latent dynamical system. Existing EHR foundation models provide powerful representations of longitudinal records, and autoregressive EHR foundation models can produce time-specific representations or generate trajectories by conditioning on observed event prefixes. Our distinction is therefore not that existing models are incapable of temporal prediction or simulation, but rather in the modeling objective: instead of representing patient evolution implicitly through autoregressive event prediction, we explicitly learn an action-conditioned transition operator over a time-indexed latent patient state. This formulation separates endogenous patient state from exogenous clinical actions and provides a representation whose primary object is the trajectory itself rather than a pooled summary of the record.

At the core of the model is a state-dependent linearization operator that propagates local changes in latent patient state while conditioning on observed clinical actions. This construction provides a lightweight inductive bias for incremental physiological evolution and supports multi-step rollout without directly regressing each future state independently. Importantly, the resulting intervention-conditioned trajectories should not be interpreted as causal estimates of treatment effects; rather, they characterize how the learned observational dynamics respond to alternative action sequences and provide a substrate for future decision-aware modeling under appropriate causal assumptions.

We evaluate the learned dynamics at multiple levels. First, we measure latent trajectory fidelity by comparing predicted rollouts with encoder-derived trajectories and assess whether temporal ordering and task-relevant structure are preserved. Second, we decode free-running latent rollouts back into the original clinical feature space and evaluate forecasting accuracy directly against observed future measurements, providing a complementary measure of trajectory fidelity beyond latent-space agreement. We compare these forecasts against task-matched temporal baselines, including regularized linear models, GRU-D, Latent ODE, and Transformer-based forecasting models, across increasing prediction horizons. We further evaluate downstream representations using classification and time-to-event tasks, with complementary metrics for class imbalance and calibration, and examine how alternative clinical action sequences perturb the learned trajectories.

Together, these experiments support a more specific claim than predictive superiority alone: a compact latent dynamics model can retain clinically relevant predictive information while exposing an explicit, time-resolved representation of patient evolution. More broadly, our results suggest that world modeling provides a useful framework for clinical machine learning in which patient histories are treated not only as records to summarize, but as trajectories whose evolution can be forecasted, interrogated, and simulated under varying observed clinical actions.

\section{Related Work}
\label{sec:related}

\textbf{World Models} Predictive world models emerged from model-based reinforcement learning, which aims to learn forward models of agent-environment dynamics \cite{ha2018world, m2023model, luo2024survey, ding2025understanding}. Early approaches based on recurrent neural networks \cite{rumelhart1985learning} focused on compressing sensory streams into predictive state representations that could be rolled forward to support planning and control \cite{guan2023leveraging, sekar2020planning}. Over time, this perspective converged with advances in deep generative modeling \cite{salakhutdinov2015learning, oussidi2018deep}, leading to architectures that learn compact spatiotemporal representations and use them as substrates for downstream decision-making. The modern formulation popularized by Ha and Schmidhuber \cite{ha2018world} demonstrated that learned latent dynamics could support policy optimization within imagined rollouts, separating representation and dynamics learning from downstream control. More broadly, this perspective treats predictive state modeling as a mechanism for learning representations sufficient to anticipate future observations and support sequential decision-making.

Another family of predictive models, Joint-Embedding Predictive Architectures (JEPA), has evolved from static representation learning toward modeling structured dynamics directly in latent space \cite{lecun2022path, brotee2025survey}. Early variants such as I-JEPA \cite{assran2023self} and subsequent extensions such as V-JEPA \cite{bardes2024revisiting, assran2025v} demonstrate that predicting latent embeddings from context can produce high-level semantic and temporal representations without explicit reconstruction. This paradigm has since been extended across a range of applications and modalities \cite{fei2023jepa, larey2026jepa, bardes2023mc, thimonier2024t, maes2026leworldmodel, luo2026toward, li2026glucofm}. More recent work, such as Causal JEPA \cite{nam2026causal}, extends this framework with intervention-aware objectives that encourage representations to distinguish changes induced by interventions from naturally occurring variation, providing a basis for reasoning over action-conditioned futures.

Recent advances such as LeWorldModel (LeWM) \cite{maes2026leworldmodel} further improve the stability of latent predictive world models through regularization mechanisms such as SIGReg \cite{balestriero2025lejepa}, enabling end-to-end learning of predictive latent dynamics without reconstructing the observation space. Our work builds directly on this predictive representation-learning framework, but specializes it to longitudinal EHR by introducing an explicit decomposition between patient state and clinical action and a state- and action-conditioned transition operator over latent increments. Thus, the contribution is not a new general-purpose world-model training framework, but a controlled latent-dynamics formulation designed for intervention-conditioned patient trajectories.

% \textbf{Predictive World Models}

\textbf{EHR foundation models and longitudinal representations:}
Foundation models for health have emerged as a rapidly growing area of research, with large-scale pretraining increasingly applied across clinical text \cite{lee2025mb, alsentzer2019publicly}, medical imaging \cite{ma2024segment, an2025raptor}, physiological signals \cite{lee2025foundation, pan2026hipno, lee2025towards, lee2025himae}, genomics \cite{larey2026jepa}, and structured data like electronic health records \cite{wornow2023shaky, gorla2025cacti, gorla2026illusion} to learn transferable representations for downstream clinical tasks. Within structured EHR, foundation modeling often adopts the language-modeling abstraction, representing medical codes and clinical events as tokens within longitudinal sequences \cite{steinberg2021language}. Early unsupervised approaches learned fixed-dimensional patient representations that compressed heterogeneous records into compact embeddings \cite{miotto2016deeppatient}. More recently, Transformer-based models have adapted BERT- and GPT-style pretraining objectives to structured EHR, producing representations that support transfer across a wide range of downstream clinical prediction tasks \cite{devlin2019bert,li2020behrt,rasmy2021medbert,odgaard2024core, an2025dk}. Subsequent approaches incorporate temporal information, visit structure, and event timing directly into the representation while retaining a general-purpose pretrained interface for downstream prediction \cite{pang2021cehrbert}.

In parallel, autoregressive EHR FM modeling has emerged as a general-purpose strategy for longitudinal EHR, using next-event or time-to-event prediction to learn transferable representations \cite{steinberg2021language,wornow2023ehrshot,steinberg2024motor,jing2026one,shmatko2025learning,zhang2025chronoformer}. These models have substantially advanced scale, transfer, and benchmarking for longitudinal EHR \cite{kolo2024meds,guo2023ehr,lee2025clinical,mcdermott2026meds}. Importantly, autoregressive EHR models are not restricted to static patient summaries: representations can be extracted at intermediate timepoints, and generative models can produce future clinical trajectories by conditioning on observed prefixes or inserted events \cite{shmatko2025learning}. The distinction we draw is therefore one of modeling objective rather than capability. Whereas autoregressive approaches primarily model the distribution of future clinical events or tokens, our formulation directly parameterizes the evolution of a latent patient state under an explicit clinical action sequence.

% Recent healthcare-specific world models and patient simulation approaches similarly move beyond static representations toward explicit modeling of longitudinal patient evolution \cite{mu2026ehrworld,lutsker2026simulating}. These works demonstrate the growing interest in generative and dynamical formulations of patient trajectories. Our formulation differs in explicitly decomposing each trajectory into an endogenous patient state and exogenous clinical actions, and learning a state- and action-conditioned transition operator over latent state changes. This decomposition makes intervention conditioning part of the transition dynamics themselves rather than representing treatments solely as additional events in an autoregressive sequence.

\begin{figure*}[t!]
    \centering
    \includegraphics[width=\linewidth]{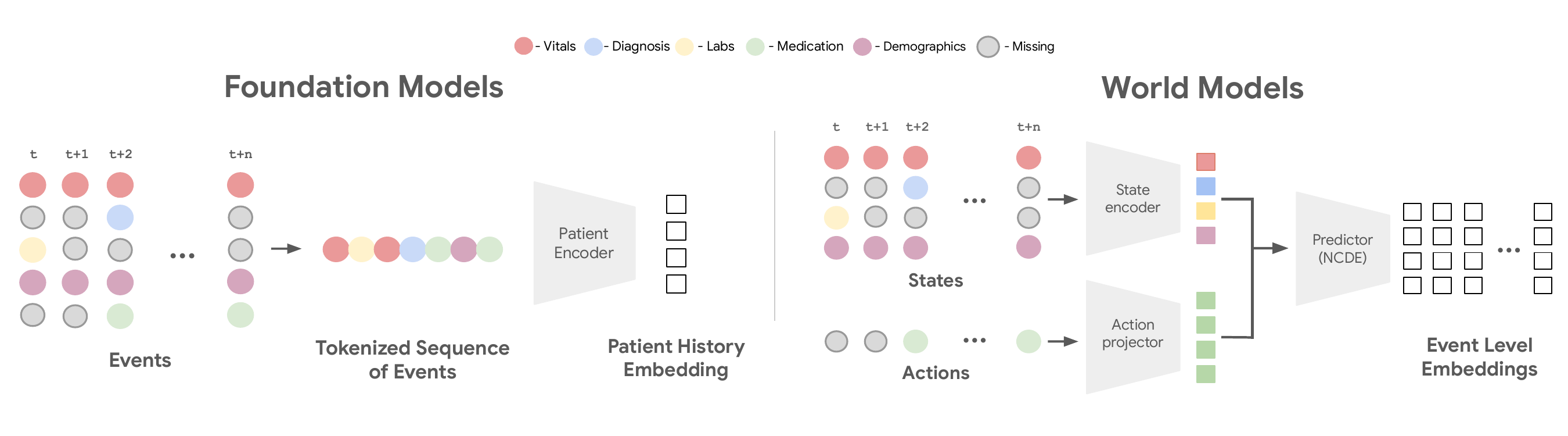}
    \caption{\small \textbf{From sequence-based EHR representations to trajectory-based world models.}
Foundation models commonly represent electronic health records as longitudinal sequences of heterogeneous events (e.g., vitals, diagnoses, labs, medications) and expose learned representations of the observed patient history for downstream prediction. In contrast, our world model formulation explicitly decomposes the record into patient states and clinical actions and models their evolution within a controlled latent trajectory system. This yields time-indexed latent states whose transitions are directly conditioned on observed clinical actions, enabling trajectory forecasting and intervention-conditioned simulation.}
    \label{fig:1}
\end{figure*}

\textbf{Our Contributions}
Building on these lines of work, we introduce a predictive world model tailored to structured EHR that represents patient histories as controlled latent trajectories. Rather than replacing autoregressive sequence modeling with another generative sequence model, we formulate patient evolution as an explicit latent dynamical system whose state is indexed over time and whose transition dynamics are conditioned on observed clinical actions (Figure \ref{fig:1}). Concretely, we introduce a state-dependent linearization transition operator, motivated by the controlled-dynamics perspective of Neural CDEs \cite{kidger2020neuralcde}, that propagates local latent increments through a state- and action-dependent linear map.

This construction provides an inductive bias toward incremental patient evolution while maintaining a lightweight transition model suitable for multi-step rollout. The resulting representation exposes the evolving latent patient state directly, enabling time-resolved prediction, rollout-based forecasting, and controlled perturbation of action sequences within a common representation space. Empirically, we evaluate this formulation through latent trajectory fidelity, forecasting in the original clinical feature space, downstream classification and time-to-event prediction, and intervention-conditioned simulation.

\section{Methods}

\paragraph{Cohort construction and temporal discretization}
We construct longitudinal patient trajectories from MIMIC-IV \cite{johnson2023mimic} using the MEDS schema \cite{arnrich2024medical, mcdermott2025meds} and partition patients into disjoint training, validation, and test cohorts. To prevent information leakage, all preprocessing statistics, including normalization parameters and feature frequencies, are computed on the training split and applied unchanged to validation and test \footnote{https://github.com/Medical-Event-Data-Standard}.

The ICU provides a particularly suitable setting for piloting trajectory-based world modeling because patient state is observed and acted upon at relatively high temporal frequency. Critically ill patients undergo repeated measurement of vital signs and laboratory values, while medications and other treatments are administered throughout the admission in response to evolving clinical status. This produces comparatively dense sequences of both physiological observations and clinical actions, making the ICU a useful environment for studying whether a model can learn state transitions that remain sensitive to changes in treatment. At the same time, the ICU setting should be viewed as a controlled starting point rather than a complete representation of longitudinal healthcare, since outpatient and chronic disease trajectories are substantially sparser and unfold over much longer timescales.

We impose a discrete-time structure by partitioning each patient timeline into fixed intervals of $\Delta = 6$ hours (Experiment in Appendix Section \ref{discrete}). Within each interval, multiple observations of the same continuous variable are aggregated via the mean. Continuous features are standardized via z-score normalization computed on observed values. Each variable is then represented by a two-channel encoding consisting of its normalized value and a corresponding binary observation mask to encode missingness. For timesteps in which a feature is unobserved, its normalized value is set to zero and its mask to zero, ensuring that missingness is explicitly identifiable rather than conflated with mean-valued observations. This yields a regularly sampled sequence indexed by $t \in \{1, \dots, T\}$.

\textbf{State and action representation}
At each timestep $t$, we construct a state vector $x_t$ by concatenating clinical features derived from continuous features including vital signs, laboratory measurements, and static features including demographics. These collectively represent the patient’s physiological state and clinical context. Observed medication administrations within each time window are modeled explicitly as a binary clinical action vector $a_t \in \{0,1\}^{d_a}$. The resulting sequence $\{(x_t, a_t)\}_{t=1}^T$ defines a discrete-time controlled trajectory.

\textbf{Feature space restriction}
To ensure a consistent representation across a large and heterogeneous population, we restrict the feature space to the top-$k$ most frequent variables computed on the training corpus. This selection is applied separately across continuous variables (e.g., vitals and labs) and intervention codes (e.g., medications). Limiting to a shared top-$k$ feature set mitigates high-cardinality effects and reduces sparsity, yielding a fixed-dimensional representation that is stable across patients and scalable to large cohorts.

\textbf{Sequence construction}
Each patient trajectory is represented as a sequence $\{(x_t, a_t)\}_{t=1}^T$ over discretized time, providing a structured and uniformly sampled approximation of the underlying clinical process suitable for downstream sequential modeling.

\subsection{State-Dependent Linearization World Model}

We adopt the LeWorldModel framework \cite{maes2026leworldmodel}, a latent predictive modeling paradigm inspired by JEPA-style objectives, and specialize it to the clinical setting by introducing a controlled dynamical system over patient trajectories. The central component of our formulation is a \emph{state-dependent linearization predictor}, which models clinical actions as inputs that modulate latent transition dynamics.

\textbf{Latent state construction}
Observations $x_t \in \mathbb{R}^{D_x}$ are mapped to a latent state $z_t \in \mathbb{R}^D$ via an encoder $f_\phi$,
\begin{equation}
z_t = f_\phi(x_t),
\end{equation}
where $f_\phi$ is a feedforward projection network. The learned representation $z_t$ provides a compact predictive state from which subsequent latent evolution is modeled. We use a latent dimensionality of $D=128$, following the scale of related latent-dynamics models \cite{wang2026temporal}. Appendix Section \ref{jacobian} further analyzes the effective rank and spectral structure of the learned transition operators, showing that the model utilizes a substantial fraction of the available latent dimensions.

\textbf{Controlled latent dynamics via state-dependent linearization.}
We model temporal evolution as a controlled discrete-time dynamical system over latent increments. Given a latent state $z_t \in \mathbb{R}^D$ and clinical action $a_t$, the predictor outputs a state- and action-dependent linear operator
\begin{equation}
F_\theta(z_t, a_t) \in \mathbb{R}^{D \times D},
\end{equation}
which parameterizes the local propagation of latent changes.

Let $\Delta z_t = z_t - z_{t-1}$. The next state is predicted as
\begin{equation}
\hat z_{t+1} = z_t + F_\theta(z_t, a_t)\,\Delta z_t,
\end{equation}
which propagates the previous latent displacement through a state- and action-dependent linear map. Since $\Delta z_1$ depends on the unobserved $z_0$, we initialize the dynamics at $t=2$ and apply the update for $t \geq 2$.

This parameterization is motivated by local linearization of smooth dynamical systems. For a latent transition map $G$, small perturbations around the current state can be approximated by
\begin{equation}
\delta z_{t+1}
\approx
J_z G(z_t,a_t)\,\delta z_t,
\end{equation}
where $J_z G(z_t,a_t)$ denotes the Jacobian of the transition with respect to the latent state at a fixed action. We therefore learn $F_\theta(z_t,a_t)$ as a state- and action-dependent approximation to such a local transition operator, using the observed latent increment $\Delta z_t$ as the propagated displacement. Conditioning the operator on $a_t$ allows different clinical action sequences to induce different local transition dynamics.

The resulting update is multiplicative in $\Delta z_t$ and does not include a separate additive action-forcing term. Consequently, the formulation places an inductive bias on actions modulating existing latent dynamics; the associated stationary-state limitation is discussed further in the Appendix. With respect to $z_t$ alone, the recurrence is second-order because it depends on $z_{t-1}$ through $\Delta z_t$. Equivalently, it can be written as a first-order controlled system in the augmented state $(z_t,\Delta z_t)$, conditioned on $a_t$.

\textbf{Latent predictive objective}
Given a trajectory $(x_{t:t+T}, a_{t:t+T})$, the encoder produces latent states $z_t = f_\phi(x_t)$ and the predictor recursively generates $\hat z_{t+j}$ for $j \geq 1$. The model is trained using a latent prediction loss
\begin{equation}
\mathcal{L}_{\mathrm{pred}}
=
\frac{1}{k}
\sum_{j=1}^{k}
\left\|
\hat z_{t+j} - z_{t+j}
\right\|_2^2,
\end{equation}
which enforces consistency between predicted and encoder-derived future states without requiring reconstruction of the observation space. This objective encourages the latent representation and transition model to support predictable multi-step evolution from the current augmented latent state under the observed sequence of clinical actions.

To regularize the geometry of the latent space, LeWorldModel adopts Sketched Isotropic Gaussian Regularization (SIGReg) \cite{balestriero2025lejepa}, enforcing that the empirical latent distribution matches $\mathcal{N}(0, I)$ via characteristic function matching. The full objective is
\begin{equation}
\mathcal{L}
=
\mathcal{L}_{\mathrm{pred}}
+
\lambda_{\mathrm{sig}}
\mathcal{L}_{\mathrm{SIGReg}}.
\end{equation}

\textbf{Curriculum learning and training dynamics}
Learning long-horizon controlled dynamics directly can be unstable due to compounding prediction error, particularly in noisy and partially observed clinical trajectories. We therefore employ a curriculum over prediction horizon and input distribution, combining teacher forcing with a sliding-window training scheme \cite{lamb2016professor, bengio2009curriculum, zhou2026physiology}. At training time, the predictor receives either the encoder-derived latent $z_t$ or its own prediction $\hat z_t$, sampled with probability $\rho$ and $1-\rho$, respectively. This defines a stochastic rollout state
\begin{equation}
\tilde z_t =
\begin{cases}
z_t & \text{with probability } \rho, \\
\hat z_t & \text{with probability } 1-\rho.
\end{cases}
\end{equation}

We jointly increase the prediction horizon $k$ while annealing $\rho$, transitioning from fully teacher-forced one-step prediction to partially autoregressive multi-step rollouts. Training is performed over sliding windows of trajectories, allowing the model to observe diverse local contexts while gradually extending to longer effective horizons. This curriculum first emphasizes accurate local transitions and subsequently trains stable composition of the learned transition operators over longer rollouts. All optimization settings and training hyperparameters are described in Appendix \ref{optim}.

\section{Experiments}

\textbf{Trajectory fidelity and latent generative validity}
We first evaluate the extent to which the learned world model preserves the structure and task-relevant information of encoder-derived patient trajectories during autoregressive rollout. Here, the central question is whether predicted latent trajectories $\{\hat z_t\}$ generated by the world model remain aligned with the encoder-derived trajectories $\{z_t\}$ that define the model's latent prediction target. We emphasize that this analysis measures \emph{latent trajectory fidelity}: agreement between $\hat z_t$ and $z_t$ does not by itself establish fidelity to the underlying clinical observations. We therefore complement these analyses with direct decoding and forecasting in the original clinical feature space in the subsequent section.

Within the latent space, we evaluate rollout fidelity along several complementary axes, including preservation of downstream predictive signal, temporal ordering, and trajectory-level distributional structure. Rather than treating any single metric as evidence of exact patient reconstruction, these analyses test whether free-running rollouts preserve properties of the encoder-derived trajectories despite compounding prediction error.

\begin{wrapfigure}{r}{0.6\textwidth}
\centering 
\includegraphics[width=\linewidth]{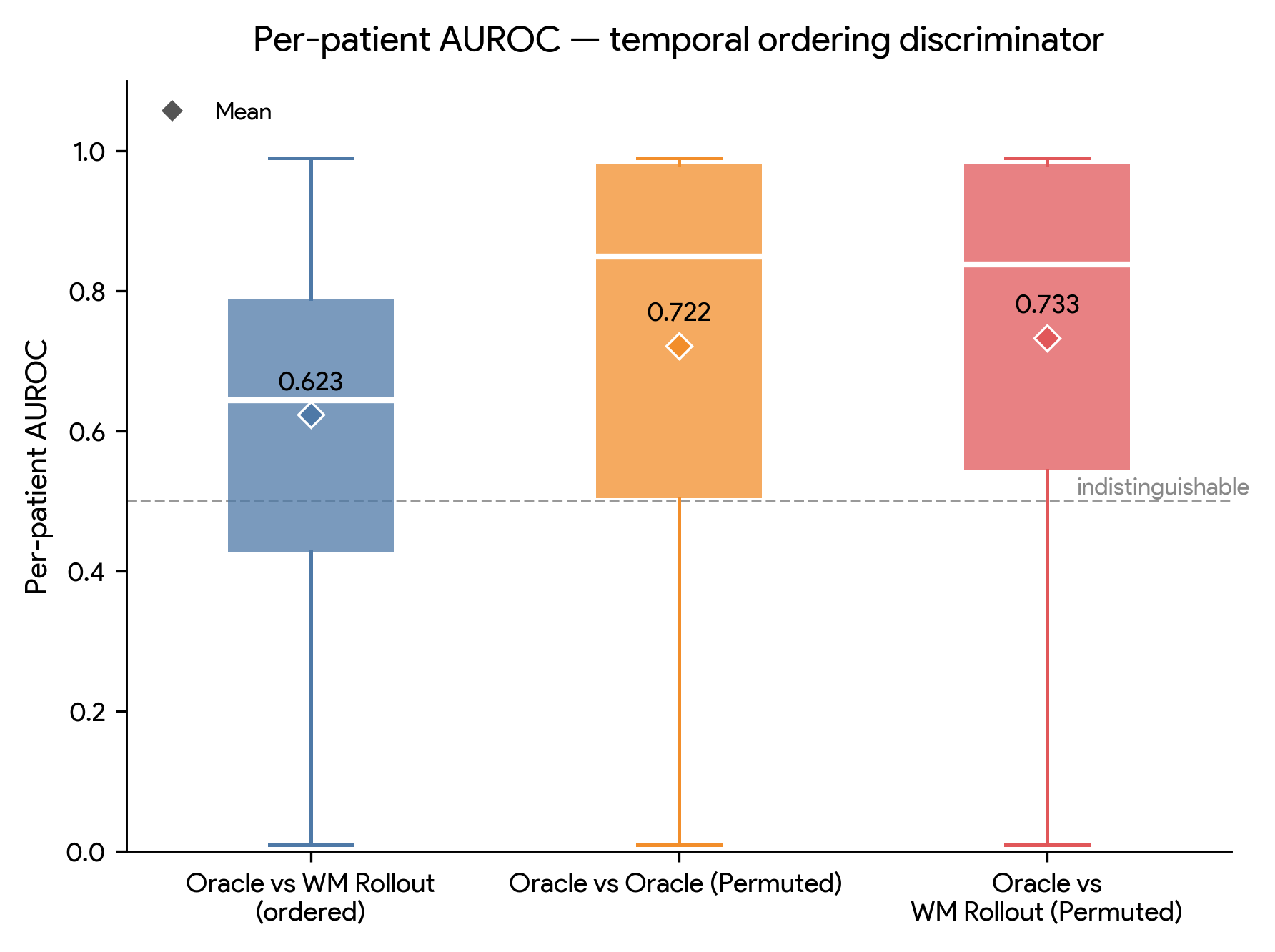}
\vspace{-8pt}
\caption{\small \textbf{Discriminability of encoder-derived and WM-generated latent trajectories.} A discriminator is trained to distinguish encoder-derived trajectories from world-model rollouts. Performance approaching chance (AUROC $\approx 0.5$) indicates limited separability in latent space. Temporal permutation increases discriminability, providing a complementary test that the learned representations contain non-exchangeable temporal structure.}
\label{gan}
\end{wrapfigure}

\textbf{Latent Feature Fidelity:} We first consider a suite of binary classification and time-to-event (TTE) prediction tasks spanning acute conditions, chronic diseases, and utilization outcomes \cite{wornow2023ehrshot}. Task definitions, prediction horizons, and cohort construction are provided in Appendix Section \ref{downstream-tasks}. For each task, we apply the same downstream probe independently to encoder-derived latent states $z_t$, denoted \textsc{Oracle}, and model-generated latent states $\hat z_t$, denoted \textsc{Rollout}. Probes are evaluated at each valid prediction timepoint along the patient trajectory and performance is averaged across timepoints using the same protocol for Oracle and Rollout representations. This paired evaluation isolates the loss of task-relevant information attributable to autoregressive rollout: if the learned predictor closely preserves the encoder-derived representation, predictive performance should remain similar between the two trajectories.

The results, summarized in Table~\ref{full-table}, show that \textsc{Rollout} and \textsc{Oracle} representations generally retain similar downstream predictive performance across the evaluated tasks, although the magnitude and direction of the difference are task-dependent. 
% This result should not be interpreted as evidence that generated trajectories exactly reproduce the underlying patient dynamics; downstream prediction can remain stable even when clinically meaningful trajectory differences are present. 
The comparison demonstrates that multi-step rollout preserves much of the task-relevant information contained in the latent representation. For highly imbalanced classification outcomes, we additionally report prevalence and AUPRC, together with calibration analyses in the Appendix, to complement AUROC-based comparisons.

\begin{figure}[t!]%{r}{0.7\textwidth}
\centering
\includegraphics[width=0.8\linewidth]{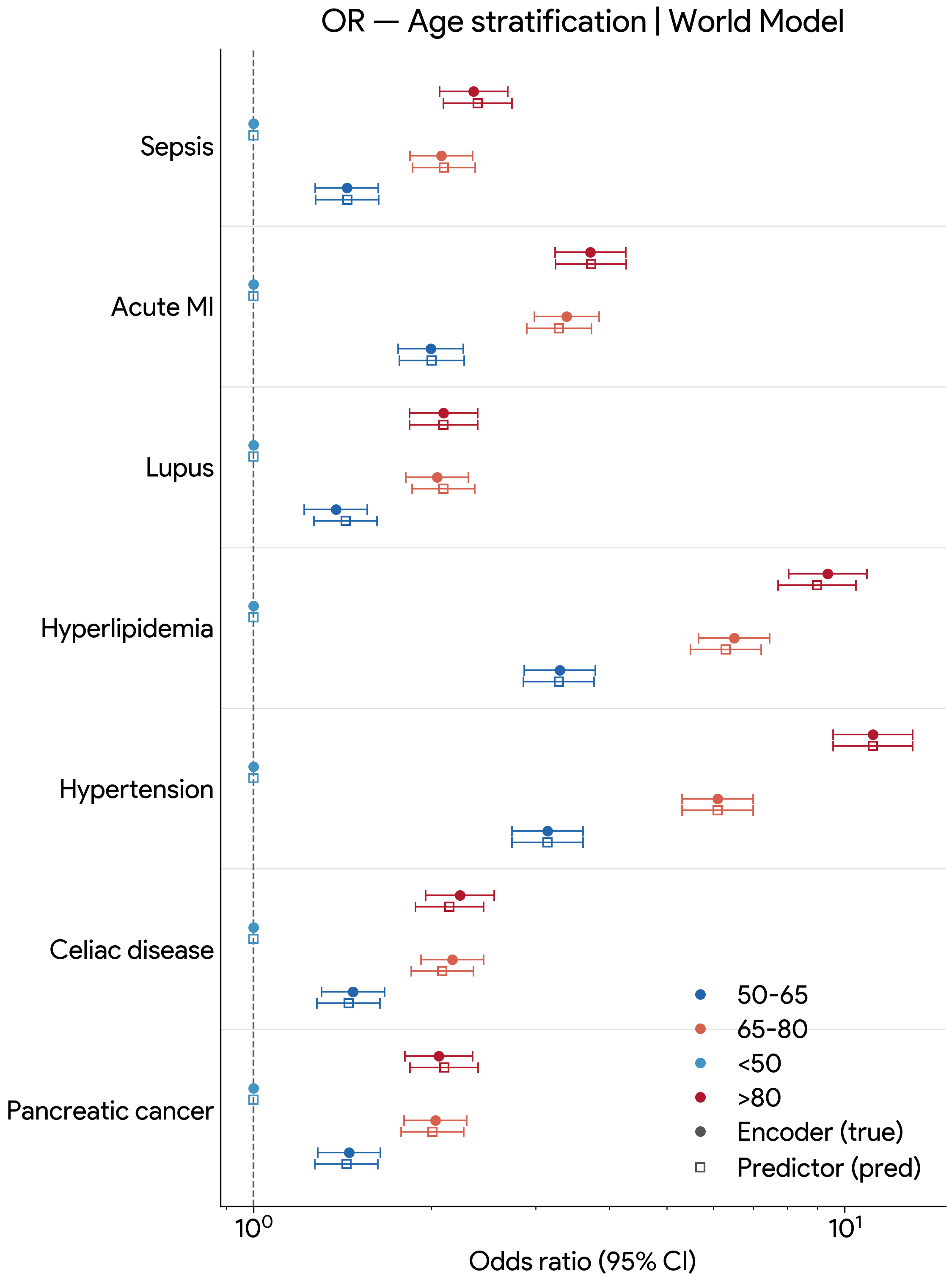}
\vspace{-8pt}
\caption{\small \textbf{Age-stratified odds ratios from generated trajectories.}
Odds ratios (95\% CI) computed from predictor-generated trajectories closely match those from encoder-derived representations across age groups, preserving relative risk structure and clinically consistent trends.}
\label{fig:age_or}
\end{figure}

\textbf{Temporal Structure Preservation} We next probe whether the learned world model preserves temporal structure using a discriminator-based evaluation between encoder-derived \textsc{Oracle} trajectories and predicted \textsc{Rollout} trajectories (Assess whether a linear classifier can tell between real and model generated trajectories). We consider three comparisons: (i) ordered \textsc{Oracle} trajectories versus ordered \textsc{Rollout} trajectories, (ii) ordered \textsc{Oracle} trajectories versus randomly permuted versions of the same trajectories, and (iii) ordered \textsc{Oracle} trajectories versus randomly permuted \textsc{Rollout} trajectories. The latter two explicitly perturb temporal order while preserving the underlying set of latent states, allowing us to test whether sequence structure itself contributes to discriminability.

As shown in Figure~\ref{gan}, discriminability is lowest when both \textsc{Oracle} and \textsc{Rollout} trajectories retain their original temporal ordering, and increases when either sequence is randomly permuted. Because the permutation operation preserves the marginal collection of latent states while disrupting their temporal arrangement, this asymmetry indicates that the learned trajectories preserve non-exchangeable temporal structure. In other words, the predictor is not merely reproducing an unordered distribution of plausible latent states; it maintains temporal dependencies that are disrupted when the sequence is shuffled. We interpret this as evidence that the rollout dynamics retain directional structure in the learned patient representation.

Lastly, we evaluate whether generated trajectories preserve clinically relevant subgroup structure by examining age-stratified odds ratios (ORs; Figure~\ref{fig:age_or}). Across the evaluated conditions, predictor-derived estimates closely track those obtained from encoder-derived representations, with substantial overlap in confidence intervals and consistent trends across age groups. Importantly, preservation of these associations under multi-step rollout indicates that the learned dynamics retain conditional risk structure rather than merely matching the marginal distribution of latent states. Known age-associated patterns, including increasing risk for several cardiovascular and metabolic conditions, remain visible in generated trajectories, suggesting that clinically relevant axes of variation are maintained during autoregressive generation. The result therefore shows that subgroup-specific statistical relationships present in the encoder-derived patient trajectories are preserved by the learned dynamics. Together with the temporal permutation and downstream prediction experiments, this provides complementary evidence that autoregressive rollout preserves structured clinical information beyond pointwise latent similarity.

\begin{figure*}%{r}{0.65\textwidth}
\vspace{-8pt}
\centering
\includegraphics[width=\linewidth]{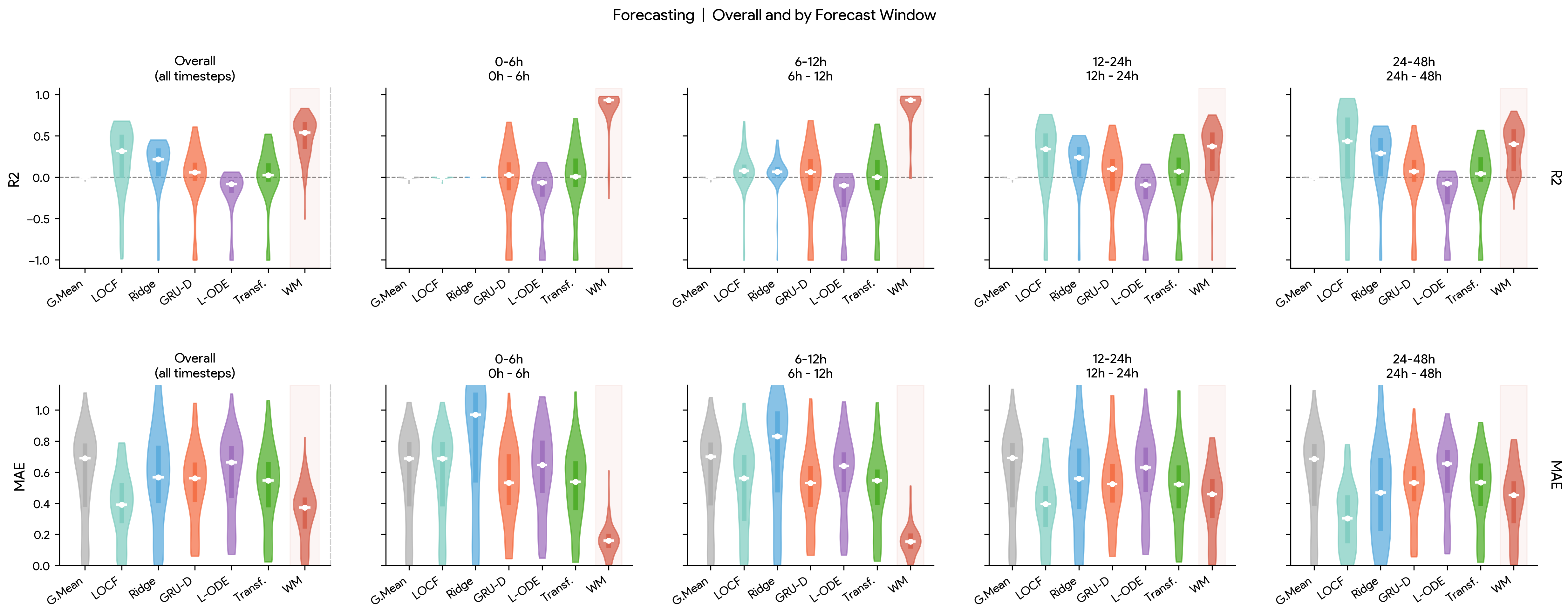}
\vspace{-10pt}
\caption{\small \textbf{Forecasting performance across prediction windows (0–6h, 6–12h, 12–24h, 24–48h) and aggregated over 79 continuous clinical variables.} We report $R^2$ (top row) and MAE (bottom row) for global mean, LOCF, Ridge Regression, GRU-D, Latent ODE, Transformer and EHR-WM baselines; violin plots denote the distribution across variables.}
\vspace{-12pt}
\label{fig:regression_by_window}
\end{figure*}

\textbf{Forecasting via rollout decoding.}
To evaluate whether EHR-WM generates trajectories that remain predictive in the original clinical feature space, we adopt a rollout-based decoding protocol that directly compares model-generated forecasts with observed future measurements. Starting from the initial observed trajectory context, we encode the patient state into latent representations and iteratively apply the learned transition dynamics to obtain future latent states $\{\hat z_{t+\tau}\}_{\tau=1}^H$. At each rollout step, a shared linear decoder maps $\hat z_{t+\tau}$ to predictions $\hat x_{t+\tau}$ over the original continuous clinical variables. After initialization, the model is rolled forward without access to ground-truth future states, such that errors in the learned transition dynamics propagate through subsequent predictions.

This evaluation complements the latent-space fidelity analyses above by measuring forecasting performance directly against observed clinical values rather than encoder-derived representations. It therefore tests whether the temporal information preserved by the latent dynamics remains recoverable in the original observation space. We compare EHR-WM against task-matched forecasting baselines, including global mean, last observation carried forward (LOCF), ridge regression, GRU-D \cite{che2016recurrent}, Latent ODE \cite{rubanova2019latent}, and a Transformer forecaster \cite{vaswani2023attentionneed}, using the same input context, prediction horizons, target variables, and evaluation protocol.

Figure~\ref{fig:regression_by_window} summarizes performance across $79$ continuous clinical variables using coefficient of determination ($R^2$) and mean absolute error (MAE) over increasing forecast horizons. Each violin plot shows the distribution of per-variable performance, allowing both central tendency and variability across clinical measurements to be assessed. EHR-WM remains competitive across horizons and maintains informative forecasts as rollout depth increases, although performance varies substantially across variables and prediction windows. Short-horizon prediction is comparatively favorable for persistence-based methods such as LOCF, whereas the relative behavior of the models changes as the forecast horizon increases and persistence becomes a weaker assumption.

Because evaluation is performed on free-running multi-step rollouts, errors introduced at one step affect subsequent predictions. The resulting degradation with forecast horizon therefore reflects both uncertainty in future clinical evolution and compounding transition error. Rather than interpreting performance at any single horizon as evidence of exact trajectory reconstruction, we use the combined $R^2$ and MAE profiles to characterize how well each model preserves clinically observed variation under increasingly long autonomous rollouts. Detailed per-variable results, including stratification by horizon, are provided in Appendix Sections~\ref{fore1} and~\ref{fore2}.

These results provide a direct observation-space complement to the latent fidelity experiments: the learned trajectories not only preserve information within the encoder representation, but also support recovery of future clinical measurements under multi-step rollout. T

\begin{table*}[t!]
\centering
\caption{\small \textbf{Comparison across classification and survival tasks with prevalence and WM rollout/oracle decomposition.}
We augment the evaluation by reporting task prevalence and decomposing EHR-WM into rollout (predicted trajectory) and oracle (ground-truth trajectory) variants, isolating representation quality from rollout error.}
\begin{adjustbox}{max width=0.95\textwidth}
\begin{tabular}{lccccccccccc}
\toprule
 & & \multicolumn{5}{c}{Binary Classification (AUC)} & \multicolumn{5}{c}{Time-to-Event (C-Index)} \\
\cmidrule(lr){3-7} \cmidrule(lr){8-12}
Task & Prev & \textsc{GBM} & \textsc{CLMBR} & \textsc{MOTOR} & \multicolumn{2}{c}{\textsc{EHR WM}} & \textsc{CoxPH} & \textsc{DeepHit} & \textsc{DeepSurv} & \multicolumn{2}{c}{\textsc{EHR WM}} \\ \cmidrule(lr){3-5}\cmidrule(lr){6-7}\cmidrule(lr){8-10}\cmidrule(lr){11-12}
 &  &  &  &  & Rollout & Oracle &  &  &  & Rollout & Oracle \\
Param Size &  & -- & 141M & 143M & \multicolumn{2}{c}{4.39M} & -- & 145k & 133k & \multicolumn{2}{c}{4.39M} \\
\midrule
Acute MI & 4.9\% & 0.723$_{\pm 0.015}$ & 0.726$_{\pm 0.013}$ & \underline{0.739$_{\pm 0.014}$} & \textbf{0.744$_{\pm 0.012}$} & {0.756$_{\pm 0.015}$} & \textbf{0.744$_{\pm 0.013}$} & 0.707$_{\pm 0.016}$ & 0.536$_{\pm 0.018}$ & \underline{0.733$_{\pm 0.013}$} & {0.746$_{\pm 0.014}$} \\
Lupus & 0.4\% & 0.711$_{\pm 0.022}$ & \underline{0.741$_{\pm 0.018}$} & \textbf{0.755$_{\pm 0.017}$} & 0.631$_{\pm 0.019}$ & 0.620$_{\pm 0.020}$ & 0.574$_{\pm 0.021}$ & \textbf{0.583$_{\pm 0.018}$} & 0.513$_{\pm 0.023}$ & \textbf{0.630$_{\pm 0.017}$} & 0.620$_{\pm 0.018}$ \\
Hyperlipidemia & 30.9\% & {0.666$_{\pm 0.010}$} & \underline{0.671$_{\pm 0.009}$} & 0.673$_{\pm 0.010}$ & \textbf{0.678$_{\pm 0.008}$} & {0.680$_{\pm 0.009}$} & 0.560$_{\pm 0.011}$ & 0.557$_{\pm 0.010}$ & \underline{0.579$_{\pm 0.009}$} & \textbf{0.624$_{\pm 0.008}$} & {0.628$_{\pm 0.008}$} \\
Hypertension & 45.2\% & 0.633$_{\pm 0.009}$ & 0.711$_{\pm 0.008}$ & \underline{0.719$_{\pm 0.008}$} & \textbf{0.734$_{\pm 0.007}$} & {0.740$_{\pm 0.008}$} & \underline{0.562$_{\pm 0.010}$} & 0.541$_{\pm 0.011}$ & \textbf{0.571$_{\pm 0.008}$} & 0.623$_{\pm 0.007}$ & {0.627$_{\pm 0.007}$} \\
Celiac & 0.5\% & \textbf{0.729$_{\pm 0.021}$} & 0.554$_{\pm 0.020}$ & \underline{0.631$_{\pm 0.017}$} & 0.627$_{\pm 0.018}$ & 0.621$_{\pm 0.019}$ & 0.527$_{\pm 0.022}$ & 0.514$_{\pm 0.021}$ & \underline{0.543$_{\pm 0.018}$} & \textbf{0.629$_{\pm 0.017}$} & 0.623$_{\pm 0.017}$ \\
Pancreatic Cancer & 0.8\% & \underline{0.811$_{\pm 0.019}$} & 0.809$_{\pm 0.018}$ & \textbf{0.831$_{\pm 0.017}$} & 0.740$_{\pm 0.020}$ & 0.749$_{\pm 0.019}$ & \textbf{0.863$_{\pm 0.015}$} & 0.725$_{\pm 0.018}$ & 0.599$_{\pm 0.021}$ & \underline{0.740$_{\pm 0.017}$} & {0.750$_{\pm 0.016}$} \\
\midrule
ICU Admission & 20.9\% & 0.771$_{\pm 0.011}$ & 0.831$_{\pm 0.010}$ & \underline{0.865$_{\pm 0.009}$} & \textbf{0.891$_{\pm 0.008}$} & {0.894$_{\pm 0.009}$} & \underline{0.748$_{\pm 0.011}$} & 0.650$_{\pm 0.013}$ & 0.594$_{\pm 0.014}$ & \textbf{0.828$_{\pm 0.009}$} & {0.827$_{\pm 0.009}$} \\
Long LOS & 15.0\% & 0.777$_{\pm 0.012}$ & \underline{0.809$_{\pm 0.011}$} & \textbf{0.821$_{\pm 0.010}$} & 0.768$_{\pm 0.013}$ & 0.746$_{\pm 0.014}$ & \underline{0.407$_{\pm 0.017}$} & 0.501$_{\pm 0.015}$ & \textbf{0.596$_{\pm 0.012}$} & 0.385$_{\pm 0.018}$ & 0.389$_{\pm 0.017}$ \\
30-day Readmission & 13.2\% & 0.712$_{\pm 0.013}$ & \textbf{0.777$_{\pm 0.011}$} & \underline{0.738$_{\pm 0.012}$} & 0.622$_{\pm 0.014}$ & 0.627$_{\pm 0.014}$ & 0.721$_{\pm 0.012}$ & \underline{0.806$_{\pm 0.010}$} & 0.547$_{\pm 0.016}$ & \textbf{0.822$_{\pm 0.011}$} & 0.816$_{\pm 0.011}$ \\
Sepsis Prediction & 4.8\% & 0.621$_{\pm 0.016}$ & 0.667$_{\pm 0.014}$ & \underline{0.707$_{\pm 0.013}$} & \textbf{0.786$_{\pm 0.012}$} & 0.782$_{\pm 0.013}$ & \underline{0.766$_{\pm 0.013}$} & 0.750$_{\pm 0.014}$ & 0.709$_{\pm 0.015}$ & \textbf{0.768$_{\pm 0.012}$} & 0.763$_{\pm 0.013}$ \\
\bottomrule
\end{tabular}
\end{adjustbox}
\label{full-table}
\end{table*}

\textbf{Benchmark comparison across modeling paradigms}
We next compare EHR-WM against established baselines across binary classification and time-to-event prediction. These include traditional tabular methods (XGBoost \cite{chen2016xgboost}), large-scale EHR foundation models (CLMBR \cite{steinberg2021language}, MOTOR \cite{steinberg2024motor}), and dedicated survival models (CoxPH \cite{cox1972regression}, DeepHit \cite{lee2018deephit}, DeepSurv \cite{katzman2018deepsurv}). These models are trained under different objectives and expose different representational interfaces: XGBoost operates on engineered tabular features, CLMBR and MOTOR learn general-purpose representations from large longitudinal EHR corpora, survival models directly optimize time-to-event objectives, and EHR-WM is trained to predict action-conditioned latent trajectories. We therefore interpret this comparison as a benchmark of downstream predictive utility across modeling paradigms rather than a controlled architectural comparison.

The goal of this evaluation is not to demonstrate uniform superiority over specialized baselines. Instead, we ask whether representations learned through a trajectory-prediction objective retain sufficient information to support standard clinical prediction tasks while also exposing an explicit latent state trajectory for forecasting and action-conditioned simulation. The paired \textsc{Oracle}/\textsc{Rollout} evaluation further separates the predictive information available in the encoder-derived representation from degradation introduced during autoregressive rollout.

For representation-based methods, we use linear probes to assess how readily downstream information can be recovered from the learned representations \cite{alain2016understanding}. For EHR-WM, the probe is evaluated independently at each valid latent state along the trajectory and performance is averaged across prediction timepoints; the same procedure is applied to both \textsc{Oracle} and \textsc{Rollout} states. CLMBR and MOTOR are evaluated using their latent representations. 
% We emphasize that these protocols reflect differences in how the models are ordinarily used and should not be interpreted as establishing that EHR foundation models are restricted to static representations: both autoregressive and representation-based EHR models can produce time-specific representations when conditioned on prefixes. 
% Our distinction is that EHR-WM directly parameterizes the transition of a persistent latent patient state under observed clinical actions. Additional time-resolved analyses are provided in Appendix Section \ref{tp}.

Table~\ref{full-table} shows that performance is heterogeneous across tasks. EHR-WM is competitive with or exceeds the comparison models on several outcomes, including Acute MI, Hyperlipidemia, Hypertension, ICU admission, and Sepsis, while substantially stronger baselines remain on other tasks, including Lupus, Pancreatic Cancer, Long LOS, and 30-day readmission. We therefore do not interpret the results as evidence of uniform predictive superiority. Rather, they show that a model trained primarily to learn latent patient dynamics retains predictive information across a diverse collection of downstream tasks, with performance that depends on the endpoint and modeling paradigm.

The comparison between \textsc{Rollout} and \textsc{Oracle} provides a separate view of rollout degradation. Across most tasks, their performance remains relatively close despite multi-step generation, indicating that the predictor preserves much of the task-relevant information encoded in the underlying latent trajectory. 
% Importantly, similarity in downstream performance is not itself evidence of exact clinical trajectory fidelity; clinically meaningful deviations may occur without substantially changing a particular classifier or survival probe. 
% We therefore evaluate trajectory fidelity separately in the original clinical feature space through rollout decoding and multi-step forecasting.
For highly imbalanced classification outcomes, we additionally report prevalence, AUPRC, and calibration analyses to complement AUROC in the Appendix.

\begin{wrapfigure}{r}{0.45\textwidth}
\vspace{-8pt}
\centering
\includegraphics[width=\linewidth]{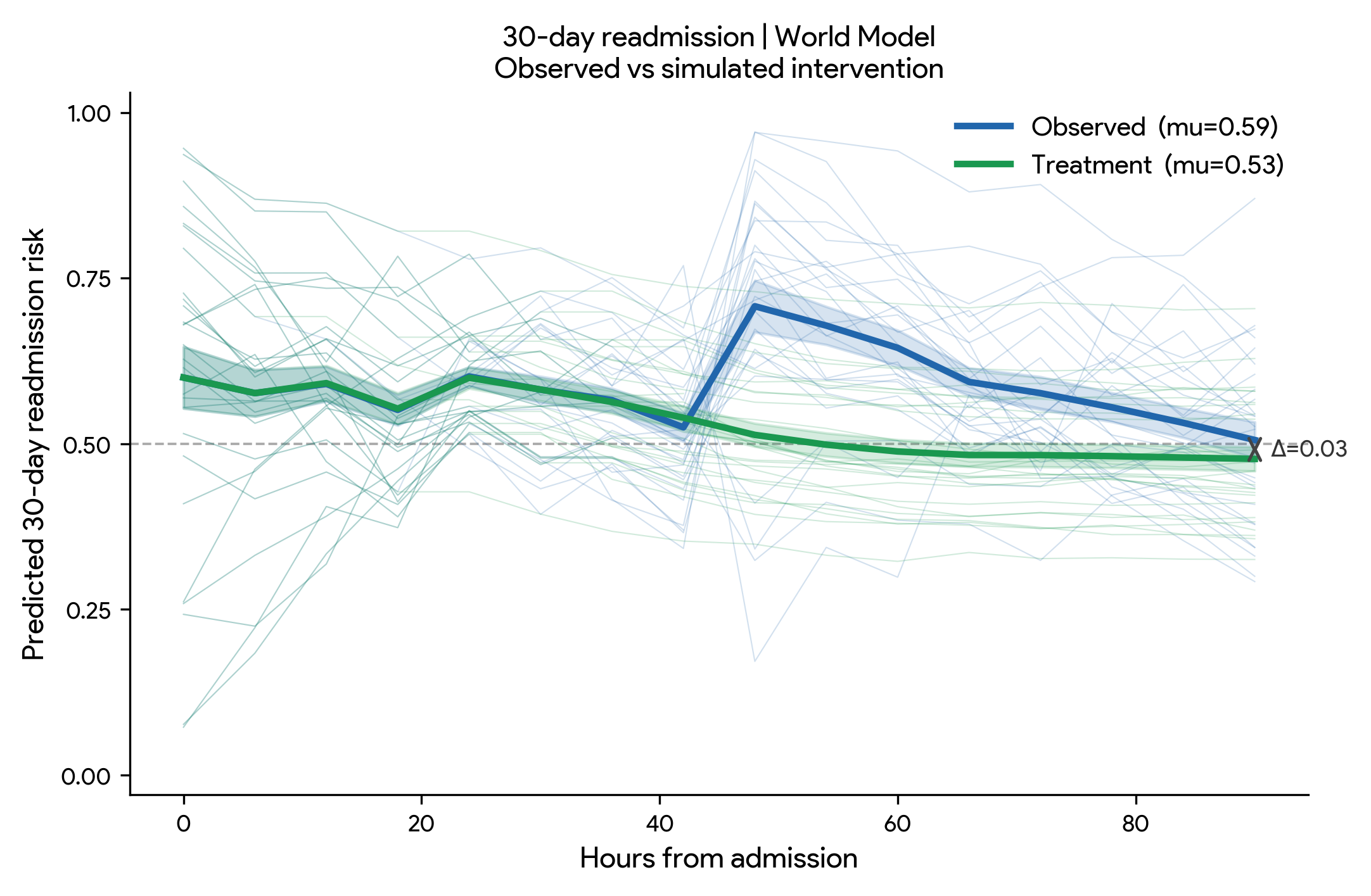}
\vspace{-10pt}
\caption{\small \textbf{Intervention-conditioned trajectory analysis for 30-day readmission.} Starting from the same initial patient state, EHR-WM generates trajectories under the observed treatment sequence (blue) and an alternative action sequence (green). Differences between the resulting risk trajectories illustrate the action dependence of the learned transition dynamics.}
\vspace{-12pt}
\label{fig:counterfactual}
\end{wrapfigure}

\textbf{Intervention-Conditioned Simulation}
The learned dynamics of EHR-WM support intervention-conditioned rollouts by propagating the same initial patient state under different clinical action sequences \cite{verma2024counterfactual}. In Figure~\ref{fig:counterfactual}, we illustrate this behavior for 30-day readmission by comparing trajectories generated under the observed treatment sequence with trajectories generated under an alternative sequence selected using model predictive control (MPC) \cite{holkar2010overview, kouvaritakis2016model, schwenzer2021review, morari1999model}. Both rollouts begin from the same latent patient state, such that differences in the resulting trajectories arise from changes to the action sequence supplied to the transition model. The alternative sequence produces a distinct time-resolved risk trajectory, demonstrating that the learned transition operator responds to changes in its clinical action input.

This experiment evaluates a structural property of EHR-WM that is distinct from conventional outcome prediction: clinical actions enter directly into the learned transition operator and can therefore modulate the subsequent latent trajectory. Rather than producing a single prediction from a fixed history, the model defines a conditional evolution process that can be queried under different action sequences. The intervention perturbation procedure, MPC objective, control horizon, and optimization setup are described in Appendix~\ref{mpc} to make the alternative-action construction explicit.

Importantly, these intervention-conditioned rollouts should not be interpreted as counterfactual or causal estimates of treatment effects. EHR-WM is trained on observational clinical data and does not identify the causal effect of an intervention from confounded treatment assignment. Instead, the experiment demonstrates that the learned dynamics are action-sensitive: holding the initial patient state fixed while modifying the supplied action sequence induces corresponding model-response changes in the generated trajectory. This capability provides a modeling interface for exploring action-conditioned patient evolution and motivates future work that combines learned world models with explicit causal assumptions or decision-making objectives.

\begin{wrapfigure}{r}{0.48\textwidth}
\vspace{-8pt}
\caption{\textbf{Ablation of training dynamics and action conditioning.}
Removing action conditioning reduces performance on selected tasks, while omitting teacher forcing, curriculum learning, or SIGReg leads to training collapse or unstable latent dynamics.}
\centering
\small
\begin{adjustbox}{max width=\linewidth}
\begin{tabular}{lccc}
\toprule
Model & ICU Adm. & Long LOS & Sepsis \\
\midrule
Full Model & 0.891 & 0.768 & 0.786 \\
w/o Action Cond. & 0.801 & 0.706 & 0.789 \\
w/o Teacher Forcing & \xmark & \xmark & \xmark \\
w/o Curriculum &\xmark & \xmark & \xmark \\
w/o SIGReg &\xmark & \xmark & \xmark \\
\bottomrule
\end{tabular}
\end{adjustbox}
\vspace{-10pt}
\label{ablation}
\end{wrapfigure}

\textbf{Ablations} Lastly, we conduct ablation studies to isolate the contributions of action conditioning and the training components used to stabilize long-horizon latent prediction. As shown in Figure~\ref{ablation}, removing action conditioning degrades performance on several tasks, although the effect is task-dependent. More importantly, without conditioning on $a_t$, the transition model can only learn dynamics as a function of the patient state and its recent evolution, effectively averaging over variation associated with different observed treatment pathways. Conditioning $F_\theta$ on $a_t$ instead allows the local transition dynamics to vary with the supplied clinical actions, providing the mechanism required for the intervention-conditioned rollout and perturbation analyses described above.

Removing teacher forcing, curriculum learning, or SIGReg results in training collapse or unstable rollout behavior, consistent with known challenges in predictive world-model training \cite{ennadir2026understanding}. In noisy and partially observed clinical trajectories, training directly on long autoregressive rollouts exposes the model to compounding prediction error before reliable local transitions have been learned. Teacher forcing reduces this instability by periodically anchoring the rollout to encoder-derived latent states, while curriculum learning progressively increases the prediction horizon so that the model first learns short-range transitions before being exposed to longer free-running trajectories. SIGReg provides a complementary role by regularizing the geometry of the learned representation and discouraging degenerate latent solutions; related variance-preserving objectives such as VICReg have similarly been used to prevent representational collapse \cite{bardes2021vicreg}. In our ablations, removing these stabilization mechanisms causes the latent representation or rollout dynamics to degenerate, indicating that successful long-horizon training depends not only on the transition architecture but also on maintaining a well-conditioned predictive latent space.

\section{Discussion}

This work reframes EHR modeling as learning a action-conditioned latent trajectory, with patient states evolving over time under observed intervention. Rather than treating this as a replacement for autoregressive EHR models, we view EHR-WM as a complementary formulation whose primary object is an explicit latent state process. Across latent-space evaluations, downstream tasks, and decoded forecasting in the original clinical feature space, the model preserves substantial predictive and temporal structure under multi-step rollout. Importantly, the raw-space forecasting results provide a stronger validation than latent agreement alone, showing that generated trajectories remain connected to observed future clinical measurements.

Performance is heterogeneous across tasks rather than uniformly superior. EHR-WM is competitive on several classification and time-to-event outcomes while specialized baselines remain stronger on others. Its main advantage is therefore not universal predictive dominance, but the combination of useful downstream representations with an explicit trajectory interface that supports forecasting and action-conditioned simulation. Similarly, changing the supplied clinical action sequence alters subsequent rollouts, demonstrating action-sensitive dynamics.

\textbf{Limitations and Future Work}
Several limitations remain. First, evaluation is restricted to MIMIC-IV, a single predominantly acute-care setting, so external validation is needed to establish generalization across institutions and more longitudinal patient populations. Second, EHR-WM operates on a restricted set of high-frequency features, which improves tractability but may omit clinically important long-tail information and partly contributes to its smaller parameter footprint relative to foundation models. 

The multiplicative transition rule also imposes a specific inductive bias: actions modulate existing latent increments rather than introducing an independent state change. In particular, when $\Delta z_t \approx 0$, the model has limited ability to produce an action-driven displacement. Future work could explore additive action terms or more flexible transition operators. More broadly, the appropriate latent dimensionality remains unresolved; adaptive, sparse, or hierarchical representations \cite{lee2025himae} may provide a better match to the heterogeneity of longitudinal EHR.

% \paragraph{Acknowledgements}

% Simon A. Lee is supported by the Warren Alpert Fellowship. We thank the Warren Alpert Foundation for their generous support of this project, Warren Alpert-UCLA Computational Biology/AI (CBAI) Scholar Training and Retention Program (award 20242540). We declare no competing interests.

\bibliography{main}

%%%%%%%%%%%%%%%%%%%%%%%%%%%%%%%%%%%%%%%%%%%%%%%%%%%%%%%%%%%

\newpage
\appendix
\section*{Appendix}
% \tableofcontents

\section{Broader Impacts}
\label{bi}

The introduction of predictive world models for Electronic Health Records (EHR) represents a significant shift toward intervention-aware clinical decision support, moving beyond static risk scores to simulate potential patient futures under various treatment sequences. By explicitly disentangling endogenous physiological progression from exogenous clinical actions, this framework provides a principled foundation for personalized medicine in complex care settings. Furthermore, the model’s high parameter efficiency—achieving competitive performance with 30× fewer parameters than standard foundation models, lowers the computational barrier for deploying advanced AI in resource-constrained clinical environments. However, the reliance on historical data necessitates careful oversight to prevent the propagation of systemic clinical biases, and the use of such models in practice must prioritize transparency and physician-in-the-loop validation to ensure ethical and equitable patient outcomes.

\section{Extended Related Works}

\textbf{Neural differential equations and controlled dynamics: }
Neural differential equation models place continuous-time dynamics at the core of representation learning by parameterizing a vector field with a neural network and integrating it with a numerical solver. The canonical Neural ODE formulation treats depth (or time) as continuous, yielding a flexible continuous-time analogue of residual networks and enabling adaptive computation through solver tolerances \cite{chen2018neuralode}. For irregularly sampled sequences, this perspective has motivated latent-variable and hybrid formulations in which latent states evolve continuously between observations and are updated at observation times, allowing principled handling of non-uniform gaps and missingness \cite{rubanova2019latentode,debrouwer2019gruodebayes}.

A key limitation of autonomous ODE dynamics in the EHR setting is semantic mismatch: an ODE trajectory is determined by its initial condition and the autonomous flow, so the latent state may drift even in the absence of new clinical evidence. Controlled differential equations address this by defining state evolution as a functional of a \emph{driving path} (the control), so that state changes are induced by observed changes in the input signal. Neural Controlled Differential Equations (Neural CDEs) operationalize this idea for irregular multivariate time series by learning a state-conditioned linear map from infinitesimal control increments to infinitesimal state increments, yielding a model that is naturally aligned with event-driven data collection \cite{kidger2020neuralcde}. Our work inherits this controlled-dynamics viewpoint but uses it to reframe EHR foundation modeling: rather than treating continuous-time modeling as a technical fix for irregular sampling, we treat it as the right object of representation learning, a population-level geometry of \emph{health-state transitions}.

\newpage

\section{Datasets}

\begin{table}[ht]
\centering
\caption{Dataset Split of MIMIC-IV}
\label{tab:dataset_split_details}
\begin{tabular}{cccc}
\toprule
\textbf{Split Name} & {Patients Count} & {Event Count} & {Number of Transitions} \\
\midrule
Training Set   & 239,770 & 424,391,148  & 6,370,510 \\
Tuning Set     & 29,971  & 51,981,308    & 787,619 \\
Test Set       & 29,871  & 52,537,879    & 791,916 \\
\bottomrule
\end{tabular}
\end{table}

\begin{table}[h!]
\centering
\caption{Patient demographics by data split (full MIMIC-IV)}
\label{tab:demographics}
\begin{tabular}{lccc}
\toprule
\textbf{Variable} & \textbf{Train} & \textbf{Val} & \textbf{Test} \\
\midrule

N patients & 239,770 & 29,971 & 29,971 \\
N encounters & 345,348 & 42,836 & 43,047 \\

\midrule
\multicolumn{4}{l}{\textit{Sex, n (\%)}} \\
Male & 113,035 (47.1\%) & 14,038 (46.8\%) & 14,086 (47.0\%) \\
Female & 126,735 (52.9\%) & 15,933 (53.2\%) & 15,885 (53.0\%) \\
Unknown & 0 (0.0\%) & 0 (0.0\%) & 0 (0.0\%) \\

\midrule
\multicolumn{4}{l}{\textit{Age at admission, years}} \\
Mean (SD) & 56.2 (20.2) & 56.5 (20.2) & 56.3 (20.2) \\
Median [IQR] & 57.6 [39.5--72.0] & 57.9 [39.7--72.4] & 57.6 [39.9--72.2] \\

\midrule
\multicolumn{4}{l}{\textit{BMI (kg/m$^2$)}} \\
Mean (SD) & 28.4 (6.8) & 28.3 (6.7) & 28.4 (6.8) \\
Median [IQR] & 27.2 [23.7--31.7] & 27.2 [23.7--31.5] & 27.2 [23.7--31.7] \\

\midrule
\multicolumn{4}{l}{\textit{Race / ethnicity, n (\%)}} \\
White & 96,347 (40.2\%) & 12,043 (40.2\%) & 12,064 (40.3\%) \\
Black / African American & 17,779 (7.4\%) & 2,241 (7.5\%) & 2,170 (7.2\%) \\
Hispanic / Latino & 7,566 (3.2\%) & 907 (3.0\%) & 935 (3.1\%) \\
Asian & 5,471 (2.3\%) & 698 (2.3\%) & 709 (2.4\%) \\
Other & 6,531 (2.7\%) & 858 (2.9\%) & 914 (3.0\%) \\
Unknown & 106,076 (44.2\%) & 13,224 (44.1\%) & 13,179 (44.0\%) \\

\bottomrule
\end{tabular}
\end{table}

We construct longitudinal patient trajectories from MIMIC-IV \cite{johnson2023mimic} in the MEDS format \cite{mcdermott2025meds, arnrich2024medical} and partition patients into disjoint training, validation, and test splits at the patient level to prevent information leakage \footnote{https://github.com/Medical-Event-Data-Standard/MIMIC\_IV\_MEDS}. The resulting dataset consists of approximately 240K patients in the training set and 30K patients each in validation and test, corresponding to over 424M clinical events and 6.3M transitions in the training split. As shown in Table~\ref{tab:demographics}, demographic characteristics are well balanced across splits, with consistent distributions in sex, age, BMI, and race/ethnicity.

\newpage

\section{Downstream Tasks}
\label{downstream-tasks}

All downstream tasks were adopted from prior benchmarks namely EHRShot \cite{wornow2023ehrshot} (classification), FoMoH \cite{pang2025fomoh} (regression/forecasting) and MEDS-DEV \cite{kolo2024meds} (classification) where cohorts were readily available via tools \cite{xu2024aces} or prior works \cite{wornow2023ehrshot}.  

\subsection{Classification Evaluation}

The goal of the classification evaluation is to predict future outcomes using only information available at each timepoint.

For disease incidence, we define a binary label over a 365-day horizon. A patient is positive if their first ICD code for a condition appears within one year after admission. Patients with no such code in that window are labeled negative. Those with a prior diagnosis before admission are excluded to ensure the model predicts new onset rather than existing disease.

We also include three operational outcomes tied to the hospital stay. Long length of stay is defined as a stay of at least 7 days. ICU transfer indicates whether any ICU-related event occurs during the admission. Thirty-day readmission indicates whether a new admission occurs within 30 days after discharge.

Patient representations are obtained by applying linear probes across the full trajectory and averaging the performance. We do this separately for embeddings derived from observed data and from predicted trajectories. For each task, we train a logistic regression classifier with standard regularization and balanced class weights. Models are evaluated on a held-out test set using AUROC, with bootstrap confidence intervals. Patients with missing or inapplicable labels are excluded per task.

\subsection{Time-to-Event (Survival) Evaluation}

We evaluate time-to-event prediction using a standard right-censored survival setup. Each patient contributes a duration, an event indicator, and a representation. The duration measures time to the event or censoring, depending on whether the outcome occurs within the observation window.

For disease incidence, duration is time from admission to first diagnosis or censoring at one year. Patients without a diagnosis are censored. For long length of stay, duration is simply the observed stay length, with events defined by stays of at least 7 days. For ICU transfer, duration is time from admission to ICU event or discharge. For readmission, duration is time from discharge to next admission, capped at 30 days, with censoring if no readmission occurs.

\subsection{Forecasting}

We evaluate forecasting capability via a linear decoding analysis on the learned patient embeddings. For each continuous target feature, we fit a probe on all training embeddings using only observed values, standardizing the embedding space before training. We then apply the same probe to held-out test embeddings and assess prediction quality separately across clinically meaningful time buckets, namely $0\text{--}6$h, $6\text{--}12$h, $(12\text{--}24)$h, and $(24\text{--}48)$h. This setup isolates how well the latent state supports downstream prediction at different stages of the patient stay, rather than conflating representation quality with temporal distribution shift. We evaluate the models using mean absolute error (MAE) and coefficient of determination $R^2$.

We compare against a set of persistence, linear, and nonlinear temporal forecasting baselines. The {global mean} baseline predicts, for each feature, its mean value estimated from the observed training data and therefore provides a context-independent reference. {Last observation carried forward (LOCF)} instead predicts the most recently observed value of each feature, providing a strong persistence baseline for slowly varying clinical measurements. {Ridge regression} provides a regularized linear forecasting baseline, learning a linear mapping from the available patient context to each future continuous target with an $\ell_2$ penalty to control overfitting.

To compare against nonlinear sequence models, we additionally evaluate {GRU-D} \footnote{\hyperlink{https://github.com/fteufel/PyTorch-GRU-D}{https://github.com/fteufel/PyTorch-GRU-D}} \citep{che2016recurrent}, which explicitly incorporates missingness and time since observation through learned decay mechanisms and is designed for sparsely observed multivariate clinical time series. {Latent ODE} \footnote{\hyperlink{https://docs.kidger.site/diffrax/examples/latent_ode/}{https://docs.kidger.site/diffrax/examples/latent\_ode/}}\citep{rubanova2019latent} represents the observed history in a latent state whose evolution is governed by a learned continuous-time ordinary differential equation, providing a continuous-dynamics comparison to our discrete action-conditioned transition model. Finally, we include a {Transformer forecaster} \citep{vaswani2017attention}, which uses self-attention over the observed temporal context to model long-range dependencies and predict subsequent clinical measurements. All forecasting baselines receive the same input context and are evaluated over the same target variables, prediction windows, and held-out patients; performance is computed only at timepoints for which the corresponding ground-truth clinical variable is observed.

An important consideration in interpreting these comparisons is that forecasting is initialized from only the first two observed timepoints, which are required to initialize the EHR-WM dynamics through the latent increment $\Delta z_t$. This provides all models with a relatively short observation history and may disproportionately limit baselines designed to exploit longer temporal context. In particular, sequence models such as GRU-D and the Transformer, as well as Latent ODE models that infer dynamics from an observed trajectory, may benefit from richer histories over which temporal dependencies, missingness patterns, and patient-specific trends can be estimated. Consequently, performance in this setting should be interpreted as forecasting from a deliberately sparse initial context rather than as a comparison under each model's potentially optimal history length.

\subsection{Model Predictive Control via Action Perturbations.}
\label{mpc}
To evaluate whether the learned dynamics encode actionable, intervention-dependent structure, we conduct an experiment based on controlled perturbations of the action sequence. Given an observed trajectory, we first train a linear probe on the latent states induced by the original sequence to predict a downstream clinical target, ensuring that the probe reflects the predictive content of the representation under observed dynamics. We then generate alternative trajectories by perturbing the action variable while holding the initial latent state fixed and rolling out the model under the modified intervention sequence. The same probe, without retraining, is applied to these simulated trajectories. Because the probe is fixed, any change in predicted risk or outcome can be attributed solely to the effect of the action perturbation on the latent dynamics. This setup isolates the sensitivity of the learned transition operator to interventions, providing a functional test of whether the model captures coherent, action-conditioned evolution rather than merely encoding static correlations. Consistent shifts in probe outputs across perturbed trajectories indicate that the world model induces structured, controllable changes in the latent trajectory aligned with downstream predictive semantics.

\section{Model Optimization and Hyperparameters}
\label{optim}

We provide a complete specification of architectural design, optimization procedure, and training hyperparameters used across all experiments to ensure reproducibility as seen in prior works. Unless otherwise stated, all configurations are held fixed across tasks to isolate representation quality from downstream model variation and ensure comparability between classification, survival, and forecasting evaluations.

\begin{figure*}%{r}{0.6\textwidth}
\vspace{-8pt}
\centering
\includegraphics[width=0.7\linewidth]{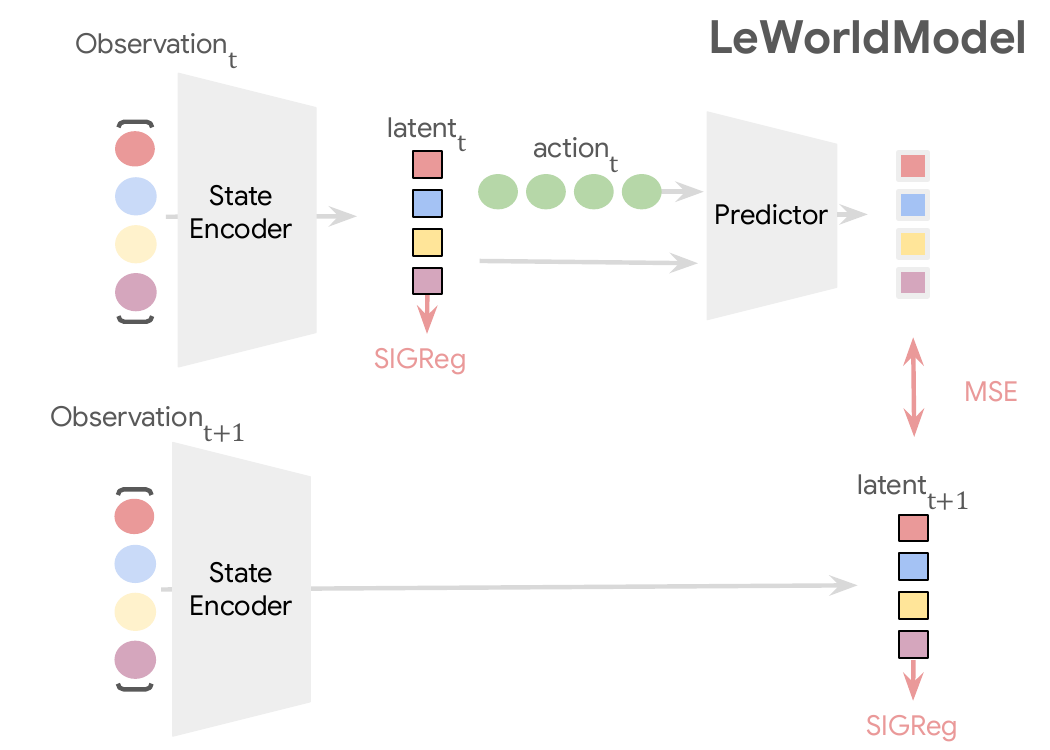}
\vspace{-10pt}
\caption{\small \textbf{Overview of the LeWorldModel framework.} At time $t$, observations are mapped into latent representations, which, together with actions, are passed to a predictor that models the subsequent latent state. The predicted latent is trained to align with the encoder-derived latent at time $t+1$ via a mean squared error objective, while structural regularization (SIGReg) is imposed on both current and future latent spaces to enforce consistency. This formulation enables learning latent dynamics directly in representation space without requiring reconstruction of the original observations.}
\label{jepa}
\end{figure*}

\begin{table*}[t]
\centering

\begin{minipage}{0.48\textwidth}
\centering
\caption{EHR WM Encoder / projector architecture.}
\label{tab:arch_encoder_projector}
\begin{adjustbox}{max width=\textwidth}
\begin{tabular}{llll}
\toprule
\textbf{Block} & \textbf{Input shape} & \textbf{Output shape} & \textbf{Details} \\
\midrule
State input $x_t$ & $D_x$ & $D_x$ & Concatenated clinical state features \\
Linear layer & $D_x$ & 256 & Dense projection to hidden space \\
LayerNorm & 256 & 256 & Stabilizes feature scale \\
ReLU & 256 & 256 & Nonlinear feature transformation \\
Dropout & 256 & 256 & $p=0.4$ regularization \\
Linear layer & 256 & 128 & Projection to latent space \\
Latent state $z_t$ & 128 & 128 & Final representation \\
\bottomrule
\end{tabular}
\end{adjustbox}
\end{minipage}
\hfill
\begin{minipage}{0.48\textwidth}
\centering
\caption{EHR WM Predictor architecture}
\label{tab:arch_predictor}
\begin{adjustbox}{max width=\textwidth}
\begin{tabular}{llll}
\toprule
\textbf{Block} & \textbf{Input shape} & \textbf{Output shape} & \textbf{Details} \\
\midrule
Latent state $z_t$ & 128 & 128 & Current latent state \\
Action input $a_t$ & $d_a$ & $d_a$ & Binary intervention vector \\
Concatenation & $128 + d_a$ & $128 + d_a$ & Joint state-action input \\
SpectralNorm Linear & $128 + d_a$ & 256 & Lipschitz-constrained mapping \\
ReLU & 256 & 256 & Nonlinearity \\
SpectralNorm Linear & 256 & $128 \times 128$ & Transition operator parameters \\
Reshape & $128 \times 128$ & $128 \times 128$ & Matrix $F_\theta(z_t, a_t)$ \\
\bottomrule
\end{tabular}
\end{adjustbox}
\end{minipage}

\end{table*}

The predictor outputs a matrix-valued function $F_\theta(z_t, a_t) \in \mathbb{R}^{D \times D}$, which defines a local linearization of the latent dynamics. Spectral normalization is applied to all predictor layers to control the operator norm of $F_\theta$, preventing instability during multi-step rollouts and implicitly constraining the Lipschitz constant of the induced dynamical system.

\subsection{Architecture}

The model instantiates a controlled latent dynamical system in which observations $x_t \in \mathbb{R}^{D_x}$ are mapped to a compact latent state $z_t \in \mathbb{R}^{D}$, and temporal evolution is governed by a state- and action-dependent linear operator. The architecture decomposes into two components: (i) an encoder/projector that defines the latent representation, and (ii) a predictor that parameterizes the transition operator.

The encoder is implemented as a shallow projection network that maps high-dimensional, heterogeneous clinical inputs into a low-dimensional latent space. The use of LayerNorm and dropout induces stability and prevents overfitting in the presence of sparse and noisy inputs. The final latent dimensionality is fixed at $D=128$, which we found to be sufficient to capture clinically meaningful variation while maintaining a compact state space.

\subsection{Loss Function}

Training is driven by a latent prediction objective over multi-step rollouts. Given predicted latent states $\hat{z}_{t+k}$ and encoded targets $z_{t+k}$, we minimize

\[
\mathcal{L}_{\text{pred}} = \frac{1}{k} \sum_{j=1}^k \| \hat{z}_{t+j} - z_{t+j} \|_2^2.
\]

To regularize the latent space, we incorporate a Sketched Isotropic Gaussian Regularization (SIGReg) term \cite{balestriero2025lejepa}, encouraging the empirical latent distribution to match $\mathcal{N}(0, I)$. The full objective is

\[
\mathcal{L} = \lambda_{\text{mse}} \mathcal{L}_{\text{pred}} + \lambda_{\text{sigreg}} \mathcal{L}_{\text{SIGReg}}.
\]

The prediction loss enforces temporal consistency in latent space, while the regularization term prevents collapse and ensures that representations remain well-conditioned for downstream linear probing.

\subsection{Optimization and Training Schedule}

We optimize all parameters using AdamW with decoupled weight decay \cite{loshchilov2017decoupled}. Let $\theta$ denote model parameters; updates take the form

\[
\theta \leftarrow \theta - \eta \left( \nabla_\theta \mathcal{L} + \lambda_{\text{wd}} \theta \right),
\]

with gradient clipping applied to enforce $\|\nabla_\theta\|_2 \leq 1.0$, stabilizing training under long rollout horizons.

Training follows a staged curriculum over rollout horizon $k$ and teacher forcing ratio $\rho$ \cite{narvekar2020curriculum}. Early stages use $\rho=1$ (fully teacher-forced), ensuring accurate local dynamics. Subsequent stages progressively decrease dynamically $\rho$ while increasing $k$, transitioning toward autoregressive multi-step prediction.

\begin{table*}[t]
\centering
\setlength{\tabcolsep}{0pt}
\renewcommand{\arraystretch}{1.0}

\begin{tabular}{@{}p{0.48\textwidth}@{\hspace{0.04\textwidth}}p{0.48\textwidth}@{}}

% ── Top-left ────────────────────────────────────────────────────────
\begin{minipage}[t]{\linewidth}
\vspace{0pt}
\centering
\caption{\small \textbf{Optimization hyperparameters.} Core training settings governing optimization dynamics, including learning rate, regularization, and gradient stabilization.}
\label{tab:optimization}
\begin{adjustbox}{max width=\linewidth}
\begin{tabular}{ll}
\toprule
\textbf{Parameter} & \textbf{Value} \\
\midrule
Optimizer & AdamW \\
Learning rate & $3 \times 10^{-4}$ \\
Weight decay & $1 \times 10^{-3}$ \\
Gradient clipping & 1.0 \\
\bottomrule
\end{tabular}
\end{adjustbox}
\end{minipage}
&
% ── Top-right ───────────────────────────────────────────────────────
\begin{minipage}[t]{\linewidth}
\vspace{0pt}
\centering
\caption{\small \textbf{Curriculum schedule.} Progressive rollout strategy controlling teacher forcing ($\rho$) and prediction horizon ($k$) across training stages.}
\label{tab:curriculum}
\begin{adjustbox}{max width=\linewidth}
\begin{tabular*}{\linewidth}{@{\extracolsep{\fill}} l c c c c}
\toprule
\textbf{Stage} & $\boldsymbol{\rho}$ & $\boldsymbol{k}$ & \textbf{Epochs} & \textbf{Patience} \\
\midrule
Stage 1 & 1.0     & 1 & 50 & 8  \\
Stage 2 & dynamic & 2 & 50 & 16 \\
Stage 3 & dynamic & 4 & 50 & 16 \\
Stage 4 & dynamic & 8 & 50 & 16 \\
\bottomrule
\end{tabular*}
\end{adjustbox}
\end{minipage}
\\[0.8em]

% ── Bottom-left ─────────────────────────────────────────────────────
\begin{minipage}[t]{\linewidth}
\vspace{0pt}
\centering
\caption{\small \textbf{Loss weighting coefficients.} Relative scaling of reconstruction and structural regularization objectives in the training loss.}
\label{tab:loss_weights}
\begin{adjustbox}{max width=\linewidth}
\begin{tabular}{lc}
\toprule
\textbf{Parameter} & \textbf{Value} \\
\midrule
$\lambda_{\text{mse}}$ & 1.0 \\
$\lambda_{\text{sigreg}}$ & 0.05 \\
\bottomrule
\end{tabular}
\end{adjustbox}
\end{minipage}
&
% ── Bottom-right ────────────────────────────────────────────────────
\begin{minipage}[t]{\linewidth}
\vspace{0pt}
\centering
\caption{\small \textbf{Data and sequence configuration.} Key preprocessing and modeling choices defining sequence length, temporal resolution, and feature selection.}
\label{tab:data_seq}
\begin{adjustbox}{max width=\linewidth}
\begin{tabular}{lc}
\toprule
\textbf{Parameter} & \textbf{Value} \\
\midrule
Sequence length & 16 \\
Batch size & 64 \\
Time bin size & 6 hours \\
Top vitals & 30 \\
Top labs & 50 \\
Top medications & 100 \\
Minimum subject frequency & 0.05 \\
Random seed & 42 \\
\bottomrule
\end{tabular}
\end{adjustbox}
\end{minipage}

\end{tabular}
\end{table*}

This curriculum can be interpreted as first learning accurate local Jacobian structure, followed by learning stable integration of the induced vector field over longer horizons.

\subsection{Data and Sequence Configuration}

Patient trajectories are discretized into fixed-length sequences of length $T=16$, with each timestep representing a 6-hour aggregation window. This yields a regularly sampled approximation of the EHR event stream.

Feature selection is applied by restricting to the top-$k$ most frequent variables within each modality (vitals, labs, medications), ensuring a consistent and dense representation across patients. Let $\mathcal{X}_t \in \mathbb{R}^{D_x}$ denote the resulting feature vector; missingness is encoded implicitly via zero-imputation after normalization.

This discretization balances temporal fidelity with computational efficiency, while the restricted feature space mitigates sparsity and stabilizes optimization in the presence of heterogeneous clinical data.

\subsection{Training Setup}All models were trained on a single Apple MacBook Pro equipped with an Apple M4 chip and 48\,GB of unified memory, using the Metal Performance Shaders (MPS) backend for hardware acceleration. This setup reflects a deliberate design constraint: the model is optimized for stability and efficiency under modest compute, rather than relying on large-scale distributed training. 

Training was performed using mixed precision where supported by MPS, with careful control of batch size and sequence length to remain within memory limits. The relatively small latent dimensionality ($D=128$) and shallow projection architecture enable efficient forward and backward passes, while spectral normalization in the predictor avoids numerical instabilities that can otherwise be amplified on non-CUDA backends.

Despite the absence of GPU clusters, the curriculum-based training procedure remained stable across all stages, with no observed divergence or collapse. Empirically, the combination of gradient clipping, teacher forcing, and staged rollout scheduling was sufficient to maintain well-conditioned optimization dynamics throughout training. This suggests that the proposed world model formulation is not only parameter-efficient but also compute-efficient, making it practical for deployment in resource-constrained research settings. Training the full model end to end took roughly 13 hours to converge.

\section{Theoretical Motivation}
\label{motiv}
\subsection{Jacobian Induced Prior}

Our predictor parameterizes a state- and action-dependent linear operator
\[
F_\theta(z_t,a_t)\in\mathbb{R}^{D\times D},
\]
which acts on latent increments and induces the update
\[
z_{t+1} = z_t + F_\theta(z_t,a_t)\Delta z_t,
\qquad \Delta z_t = z_t - z_{t-1}.
\]
This choice can be interpreted as learning a local Jacobian field over latent space. Rather than predicting the next state directly, the model learns how infinitesimal changes in the current latent state propagate under interventions. In matrix form, a first-order local linearization of the latent transition map around $(z_t,a_t)$ takes the form
\[
z_{t+1} \approx z_t + J_G(z_t,a_t)\Delta z_t,
\]
where
\[
J_G(z_t,a_t)=
\begin{bmatrix}
\frac{\partial G_1}{\partial z_1} & \cdots & \frac{\partial G_1}{\partial z_D} \\
\vdots & \ddots & \vdots \\
\frac{\partial G_D}{\partial z_1} & \cdots & \frac{\partial G_D}{\partial z_D}
\end{bmatrix}
\]
is the Jacobian of an underlying latent transition map $G$. Our model replaces this unknown Jacobian with a learned approximation $F_\theta(z_t,a_t)$.

This parameterization is useful for EHR because patient trajectories are event-driven, and locally smooth over short horizons. In such settings, a Jacobian-style model is a natural inductive bias: it encodes local sensitivity, respects the geometry of incremental changes, and allows interventions to alter the local flow without requiring a global closed-form transition. In contrast to direct next-step regression, which may treat each transition as an unrelated mapping, the Jacobian view ties all predictions to a shared dynamical structure.

A simple stability argument follows from the operator norm of the learned velocity matrix. If
\[
\|F_\theta(z_t,a_t)\|_2 \leq \gamma
\]
for all $(z_t,a_t)$, then for any latent increment $\Delta z_t$,
\[
\|z_{t+1}-z_t\|_2
=
\|F_\theta(z_t,a_t)\Delta z_t\|_2
\leq
\gamma \|\Delta z_t\|_2.
\]
Thus, when $\gamma<1$, the update is locally contractive: perturbations do not amplify, and repeated rollouts remain numerically stable. This explains why spectral normalization is not merely a regularizer but part of the modeling assumption itself: it enforces a bounded local Jacobian norm and prevents unstable latent dynamics.

The Jacobian formulation also gives a concrete interpretation of interventions. Since $F_\theta$ depends on $a_t$, two different action sequences induce different local linear operators, and hence different latent flows. This makes the model sensitive to treatment changes while remaining anchored to a single geometric object, namely the transition Jacobian field. As a result, the representation is both expressive and structured: it can represent heterogeneous clinical evolution, but only through locally coherent dynamical deformations.

The Jacobian-based parameterization provides three desirable properties for EHR modeling: local linear sensitivity to small changes, stability under multi-step rollout, and action-dependent modulation of latent dynamics. These properties make it a principled inductive bias for trajectory modeling in clinical time series.

\subsection{Failure Mode: Absorbing State}

\paragraph{Absorbing-State Failure Mode in Incremental Dynamics.}
A structural consequence of the update rule
\[
\hat{z}_{t+1} = z_t + F_\theta(z_t, a_t)\Delta z_t
\]
is the existence of an absorbing regime when the latent increment vanishes. If $\Delta z_t = 0$, then $\hat{z}_{t+1} = z_t$, and under autoregressive rollout the system remains fixed, rendering the trajectory unresponsive to future interventions. In discretized and partially observed EHR data, repeated observations within a bin can induce small or near-zero increments, making this a plausible failure mode of purely multiplicative dynamics.

\paragraph{Mitigation via Latent Regularization and Training Dynamics.}
In practice, exact absorbing states are rarely observed due to the combined effect of representation learning and regularization. In particular, isotropic regularization (e.g., SIGReg) enforces that latent representations maintain non-degenerate variance across dimensions and timesteps, discouraging collapse to constant embeddings where $z_t = z_{t-1}$. This implicitly maintains nonzero latent increments and prevents the model from converging to trivial fixed-point dynamics. Additionally, stochasticity introduced by partial observability and teacher forcing during training injects variation into the latent trajectory, further reducing the likelihood of zero-increment regimes. While these mechanisms empirically mitigate absorbing behavior, the reliance on multiplicative updates highlights a structural limitation: without an explicit additive forcing term, the model depends on the encoder to sustain temporal variation, suggesting a direction for more robust formulations of controlled latent dynamics.

\section{Extended Results}

\subsection{Jacobian and Stability Analysis}
\label{jacobian}

To better understand the learned latent dynamics, we analyze the spectrum and structure of the predictor-induced transition operators. Concretely, we extract latent trajectories from the held-out test set and evaluate the model in inference mode without gradient updates. For each patient sequence, we compute latent embeddings $z_t$ using the encoder and construct latent increments $\Delta z_t = z_t - z_{t-1}$. At each timestep, we then evaluate the predictor to obtain the state- and action-conditioned velocity matrix
\[
F_\theta(z_t, a_t) \in \mathbb{R}^{D \times D}.
\]

This yields a large collection of velocity matrices, which we analyze using singular value decomposition and eigendecomposition. These matrices correspond to the learned local linearization of the latent dynamics and provide a direct view of how perturbations propagate under the model.

In addition to the velocity matrices, we compute the full transition Jacobian
\[
J_t = \frac{\partial z_{t+1}}{\partial z_t},
\]
via automatic differentiation on a random subset of test states. This Jacobian captures both the explicit linear operator $F_\theta$ and its dependence on the current latent state, providing a more complete characterization of the nonlinear transition map
\[
z_{t+1} = z_t + F_\theta(z_t, a_t)\Delta z_t.
\]

Together, these two objects allow us to disentangle the intrinsic linear structure of the learned dynamics (through $F_\theta$) from additional expressive capacity introduced by state dependence (through $J_t$). We report statistics over both operators to assess stability, rank, and spectral structure of the learned dynamical system.

\paragraph{Velocity matrix statistics}

\begin{wraptable}{r}{0.42\textwidth}
\vspace{-8pt}
\centering
\caption{\small \textbf{Velocity matrix statistics for EHR-WM.} Summary of spectral and norm-based properties of the learned latent transition (velocity) operator, including measures of rank, conditioning, and magnitude, which together characterize the complexity, stability, and expressivity of the induced latent dynamics.}
\begin{adjustbox}{max width=\linewidth}
\begin{tabular}{lc}
\toprule
\textbf{Metric} & \textbf{EHR-WM} \\
\midrule
Effective rank & $106.68 \pm 5.83$ \\
Condition number & $4124.48 \pm 46739.23$ \\
Frobenius norm & $3.91 \pm 0.60$ \\
Spectral norm & $0.86 \pm 0.14$ \\
Nuclear norm & $38.50 \pm 7.10$ \\
\bottomrule
\end{tabular}
\end{adjustbox}
\vspace{-10pt}
\end{wraptable}

The learned velocity operators exhibit consistently high effective rank, occupying a substantial portion of the $D=128$ latent space. This suggests that the transition dynamics are not restricted to a low-dimensional subspace, but instead distribute variation broadly across latent directions. In contrast to many low-rank dynamical models, the system retains a high degree of expressivity, allowing multiple modes of evolution to coexist and interact within the representation.

At the same time, the spectral norm remains uniformly below 1, indicating that the local linearization is contractive in expectation. This implies that small perturbations in latent space do not amplify under rollout, providing a mechanism for controlling error accumulation over long horizons. Notably, this contractive behavior emerges without explicit global stability constraints, suggesting that the combination of spectral normalization and training objectives implicitly biases the system toward stable regimes.

The large variance observed in the condition number reflects the presence of small singular values, indicating that certain directions in latent space are weakly propagated. Rather than signaling instability, this reveals an anisotropic structure in the learned dynamics, where the model selectively attenuates variation along less informative directions while preserving dominant modes. This behavior is consistent with an implicit regularization effect, in which the transition operator balances expressivity with controlled contraction, yielding trajectories that are both rich and numerically stable over extended rollouts.

From a dynamical systems perspective, this combination of high effective rank and bounded operator norm places the model in a regime that is neither degenerate nor chaotic. Instead, it operates in a structured intermediate regime, where trajectories can explore a large portion of latent space while remaining well-conditioned. This provides a plausible explanation for the empirical stability observed in multi-step forecasting and rollout generation.

\begin{wraptable}{l}{0.42\textwidth}
\vspace{-8pt}
\centering
\caption{\small \textbf{Eigenvalue stability of $F_\theta$.} Spectral analysis of the learned transition operator, reporting the distribution of eigenvalue magnitudes to assess dynamical stability, contraction properties, and the absence of unstable or marginally stable modes in the latent evolution.}
\begin{adjustbox}{max width=\linewidth}
\begin{tabular}{lc}
\toprule
\textbf{Metric} & \textbf{EHR-WM} \\
\midrule
$|\lambda| < 1$ & 100.0\% \\
$|\lambda| \in [0.9, 1.1]$ & 0.0\% \\
$|\lambda| > 1$ & 0.0\% \\
$\max |\lambda|$ & 0.938 \\
\bottomrule
\end{tabular}
\end{adjustbox}
\vspace{-10pt}
\end{wraptable}

\paragraph{Eigenvalue stability}

The eigenvalue spectrum provides a direct view of stability in the learned \emph{increment dynamics}. All eigenvalues of $F_\theta$ lie strictly within the unit circle, indicating that the velocity field is contractive across sampled states. This ensures that latent \emph{increments} $\Delta z_t$ decay over time rather than amplify, preventing error explosion during multi-step rollout.

Importantly, this contractivity applies to the increment dynamics rather than the absolute latent state. Because the update defines a second-order recurrence, the augmented transition Jacobian $\mathcal{J}_t$ has eigenvalues given by those of $F_\theta$ together with an eigenvalue at $1$. As a result, the system is \emph{marginally stable} in the absolute state $z_t$: while deviations in $\Delta z_t$ decay, perturbations to $z_t$ can persist as steady-state shifts.

The absence of eigenvalues of $F_\theta$ near the unit boundary indicates that the model does not operate near instability in its velocity field; instead, it learns consistently damped incremental dynamics. This aligns with the use of spectral normalization, which constrains the operator norm and stabilizes the evolution of latent differences over time.

\paragraph{Full transition Jacobian}

Because the update depends on both $z_t$ and $z_{t-1}$, the model induces a second-order recurrence. We therefore consider the augmented state $x_t = [z_t; z_{t-1}] \in \mathbb{R}^{2D}$ and define the transition map
\begin{equation}
x_{t+1} = \Phi_t(x_t) =
\begin{bmatrix}
z_t + F_\theta(z_t,a_t)(z_t - z_{t-1}) \\
z_t
\end{bmatrix}.
\end{equation}
The full transition Jacobian is then
\begin{equation}
\mathcal{J}_t = \frac{\partial \Phi_t}{\partial x_t}
=
\begin{bmatrix}
I + F_t + R_t & -F_t \\
I & 0
\end{bmatrix},
\qquad
F_t := F_\theta(z_t,a_t),
\end{equation}
where $R_t = \frac{\partial F_t}{\partial z_t}(z_t - z_{t-1})$ captures higher-order state dependence. In a local frozen-coefficient approximation, $R_t$ is small, yielding
\begin{equation}
\mathcal{J}_t \approx
\begin{bmatrix}
I + F_t & -F_t \\
I & 0
\end{bmatrix},
\qquad
\det(\mathcal{J}_t) = \det(F_t).
\end{equation}
This augmented Jacobian characterizes the full end-to-end sensitivity of the latent transition, while the commonly used quantity $\partial z_{t+1}/\partial z_t = I + F_t + R_t$ corresponds only to a sub-block.

\begin{wraptable}{r}{0.42\textwidth}
\vspace{-8pt}
\centering
\caption{\small \textbf{Full transition Jacobian statistics.} Aggregate properties of the augmented transition Jacobian $\mathcal{J}_t$, capturing its effective dimensionality and conditioning, which reflect the stability and expressivity of the learned dynamics.}
\begin{adjustbox}{max width=\linewidth}
\begin{tabular}{lc}
\toprule
\textbf{Metric} & \textbf{EHR-WM} \\
\midrule
Effective rank & $127.00 \pm 1.12$ \\
Condition number & $4.16 \pm 11.25$ \\
Spectral norm & $1.24 \pm 0.12$ \\
\bottomrule
\end{tabular}
\end{adjustbox}
\vspace{-10pt}
\end{wraptable}

Empirically, $\mathcal{J}_t$ is nearly full-rank, indicating that the transition map preserves the dimensional structure of the latent space and avoids collapsing information into lower-dimensional manifolds. This stands in contrast to many representation learning settings where dimensional collapse can occur, and suggests that the model maintains a rich and locally invertible geometry across transitions.

The spectral norm exceeding $1$ is consistent with the residual structure of the update, which includes an identity component. While the full mapping can locally expand in certain directions, stability is governed by the learned operator $F_\theta$, which controls the incremental dynamics and prevents unbounded growth over successive steps.

Notably, the condition number remains low with relatively small variance, indicating that $\mathcal{J}_t$ is well-conditioned across the dataset. This suggests that the transition function does not exhibit extreme sensitivity to perturbations in specific directions, and that the latent dynamics vary smoothly rather than exhibiting sharp anisotropies. Together, these properties indicate that the model balances expressivity with numerical robustness, enabling stable multi-step rollouts while preserving high-dimensional structure.

\paragraph{Spectral concentration}

\begin{wraptable}{l}{0.42\textwidth}
\vspace{-8pt}
\centering
\caption{\small \textbf{Rank for cumulative energy of $F_\theta$.} Number of singular components required to capture increasing fractions of spectral energy, illustrating how variance is distributed across modes and indicating the degree of low-rank structure in the learned transition operator.}
\begin{adjustbox}{max width=\linewidth}
\begin{tabular}{lc}
\toprule
\textbf{Threshold} & \textbf{EHR-WM} \\
\midrule
50\% & 40 \\
75\% & 68 \\
90\% & 92 \\
95\% & 102 \\
99\% & 116 \\
\bottomrule
\end{tabular}
\end{adjustbox}
\vspace{-10pt}
\end{wraptable}

The cumulative singular value spectrum reveals moderate concentration of energy: approximately one-third of the singular directions explain half of the operator norm, while over 100 directions are required to capture near-total energy. This indicates that the learned dynamics are neither low-rank nor fully isotropic, but instead exhibit structured high-dimensional behavior.

These results reveal a consistent picture of the learned dynamics. The velocity matrices are contractive and stable, ensuring robust multi-step rollouts, while maintaining high effective rank and directional diversity. The full transition Jacobian remains nearly full-rank, indicating that state-dependent modulation enriches the dynamics beyond a fixed linear operator.

\subsection{GAN Discriminator over time}
\begin{wrapfigure}{r}{0.45\textwidth}
\vspace{-10pt}
\centering
\includegraphics[width=\linewidth]{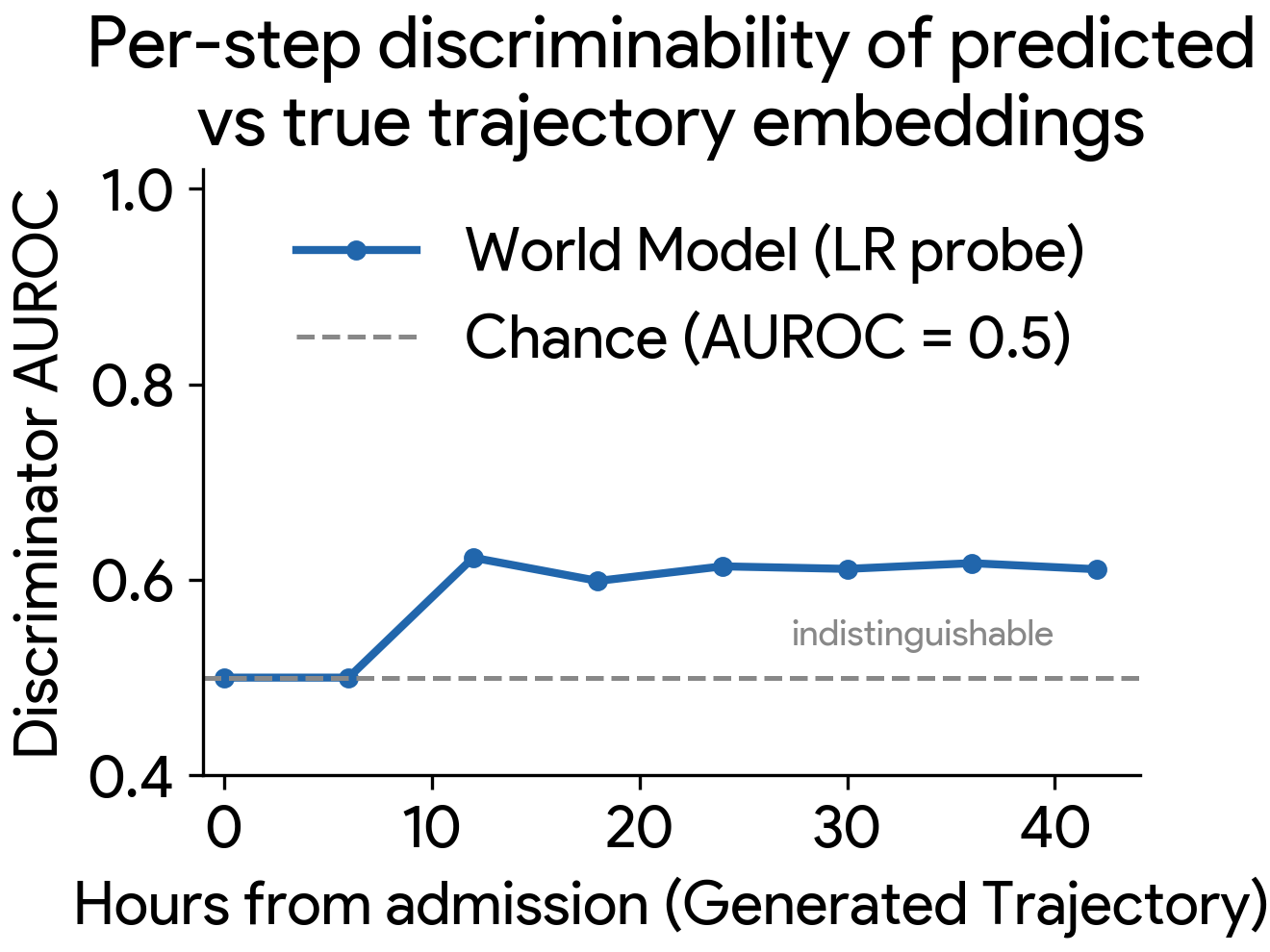}
\vspace{-8pt}
\caption{\small \textbf{Discriminability of generated vs. true trajectories.}
A linear probe is trained to distinguish predictor-generated from encoder-derived latent states at each timestep. Performance remains close to chance (AUROC $\approx 0.5$), indicating that generated trajectories are largely indistinguishable over time.}
\vspace{-25pt}
\label{gan2}
\end{wrapfigure}
To further probe generative fidelity, we adopt a discriminative evaluation protocol that operates directly in latent trajectory space (Figure~\ref{gan2}). Concretely, we construct a sequence-level classification task in which a discriminator $D_\psi$ is trained to distinguish between encoder-derived latent trajectories ${z_t}_{t=1}^T$ and predictor-generated trajectories ${\hat{z}*t}*{t=1}^T$. The discriminator receives full sequences rather than individual timesteps, thereby allowing it to exploit both marginal feature discrepancies and higher-order temporal dependencies, including autocorrelation structure and cross-time interactions that may arise from compounding prediction errors.

In parallel, we consider a stronger adversarial formulation in which $D_\psi$ is optimized in a GAN-style setup to maximally separate the two trajectory distributions, effectively probing whether any systematic statistical signal remains that could distinguish generated rollouts from true latent dynamics. Notably, across both settings, discriminator performance remains consistently near chance (AUROC $\approx 0.5$), even when evaluated per timestep and over full sequences. This suggests that the learned predictor induces a latent transition operator whose induced trajectory distribution closely matches that of the encoder, not only in terms of pointwise alignment but also in its temporal structure. Taken together, these results provide evidence that the model captures a coherent latent dynamics process rather than merely producing locally plausible but globally inconsistent predictions.

\subsection{Sensitivity to Temporal Discretization}
\label{discrete}

We ablate the temporal bin size $\Delta t$ to systematically characterize how discretization impacts representation learning and downstream predictive performance. Concretely, we evaluate four resolutions—3, 6, 9, and 12 hour bins—while holding the model architecture, optimization procedure, and training data fixed. This isolates the effect of temporal aggregation as a structural inductive bias on the learned latent dynamics. As shown in Table~\ref{tab:delta_t_ablation}, performance consistently peaks at $\Delta t = 6$ hours across all evaluated tasks, indicating that this resolution achieves a favorable balance between temporal fidelity and statistical efficiency.

\begin{wraptable}{r}{0.48\textwidth}
\vspace{-8pt}
\centering
\caption{\small \textbf{Ablation of temporal discretization}. Performance is best at $\Delta t = 6$ hours, reflecting a trade-off between sparsity at finer bins and loss of granularity at coarser bins.}
\begin{adjustbox}{max width=\linewidth}
\begin{tabular}{lccc}
\toprule
$\Delta t$ (hours) & ICU Adm. & Long LOS & Sepsis \\
\midrule
3  & 0.882 & 0.760 & 0.779 \\
6  & 0.891 & 0.768 & 0.786 \\
9  & 0.884 & 0.761 & 0.781 \\
12 & 0.876 & 0.754 & 0.774 \\
\bottomrule
\end{tabular}
\end{adjustbox}
\vspace{-10pt}
\label{tab:delta_t_ablation}
\end{wraptable}

The observed trend reflects a trade-off that can be understood through the lens of signal-to-noise and effective sample complexity. At finer granularity ($\Delta t = 3$ hours), the sequence length increases, but each bin contains fewer observed events, leading to sparsity in the input space and higher variance in the resulting representations. This fragmentation reduces the reliability of both the encoder and the learned transition dynamics, particularly for infrequent clinical measurements. Conversely, coarser discretizations ($\Delta t = 9$ and $12$ hours) aggregate over longer intervals, effectively smoothing the underlying trajectory and attenuating short-horizon variations that may carry predictive signal. This induces a bias toward overly stable dynamics and limits the model’s ability to capture rapid physiological changes.

\subsection{Classification Performance}

\begin{table*}[t!]
\centering
\caption{\small
\textbf{AUPRC comparison across binary classification tasks with
prevalence and WM rollout/oracle decomposition.}
EHR-WM rollout denotes evaluation using predicted trajectories, while EHR-WM
oracle denotes evaluation using real ICU trajectories.}
\begin{adjustbox}{max width=0.95\textwidth}
\begin{tabular}{lcccccc}
\toprule
 & & \multicolumn{5}{c}{Binary Classification (AUPRC [Approximate 95\% Interval])} \\
\cmidrule(lr){3-7}
Task & Prev & \textsc{GBM} & \textsc{CLMBR} & \textsc{MOTOR}
& \multicolumn{2}{c}{\textsc{EHR WM}} \\
\cmidrule(lr){3-5}\cmidrule(lr){6-7}
 & & & & & Rollout & Oracle \\
Param Size & & -- & 141M & 143M & \multicolumn{2}{c}{4.39M} \\
\midrule

Acute MI
& 8.6\%
& 0.191 [0.168--0.221]
& 0.194 [0.172--0.223]
& 0.208 [0.185--0.238]
& 0.214 [0.192--0.243]
& 0.205 [0.186--0.233] \\

Lupus
& 0.6\%
& 0.015 [0.011--0.023]
& 0.019 [0.015--0.027]
& 0.022 [0.018--0.029]
& 0.009 [0.006--0.015]
& 0.008 [0.006--0.012] \\

Hyperlipidemia
& 41.8\%
& 0.244 [0.225--0.260]
& 0.250 [0.231--0.265]
& 0.252 [0.233--0.268]
& 0.258 [0.240--0.273]
& 0.270 [0.252--0.284] \\

Hypertension
& 61.3\%
& 0.204 [0.292--0.218]
& 0.271 [0.259--0.285]
& 0.278 [0.266--0.292]
& 0.291 [0.279--0.204]
& 0.294 [0.281--0.206] \\

Celiac
& 0.6\%
& 0.023 [0.018--0.037]
& 0.009 [0.005--0.018]
& 0.012 [0.008--0.022]
& 0.012 [0.008--0.022]
& 0.013 [0.008--0.021] \\

Pancreatic Cancer
& 1.3\%
& 0.082 [0.065--0.113]
& 0.081 [0.064--0.111]
& 0.102 [0.084--0.134]
& 0.043 [0.029--0.068]
& 0.049 [0.033--0.078] \\

\midrule

ICU Admission
& 38.7\%
& 0.325 [0.307--0.341]
& 0.314 [0.397--0.331]
& 0.368 [0.350--0.384]
& 0.410 [0.393--0.426]
& 0.410 [0.395--0.424] \\

Long LOS
& 30.5\%
& 0.552 [0.532--0.572]
& 0.605 [0.585--0.624]
& 0.625 [0.606--0.645]
& 0.538 [0.518--0.558]
& 0.500 [0.480--0.520] \\

30-day Readmission
& 19.0\%
& 0.318 [0.305--0.335]
& 0.413 [0.399--0.430]
& 0.353 [0.340--0.369]
& 0.226 [0.214--0.241]
& 0.234 [0.220--0.249] \\

Sepsis Prediction
& 9.2\%
& 0.136 [0.109--0.163]
& 0.167 [0.139--0.195]
& 0.202 [0.174--0.232]
& 0.307 [0.275--0.340]
& 0.321 [0.291--0.354] \\

\bottomrule
\end{tabular}
\end{adjustbox}
\label{classification-auprc-table}
\end{table*}

The AUPRC results in Table~\ref{classification-auprc-table} show heterogeneous performance across tasks, with EHR-WM remaining competitive with the comparison models on several outcomes while underperforming on others. EHR-WM achieves the highest rollout AUPRC for Acute MI, ICU Admission, and Sepsis Prediction, and is also competitive for Hyperlipidemia and Hypertension. In contrast, MOTOR performs best on Lupus, Pancreatic Cancer, and Long LOS, while CLMBR achieves the strongest result for 30-day Readmission. Performance on Celiac is similar across several models, with all methods operating in a very low-prevalence regime.

The comparison between EHR-WM \textsc{Rollout} and \textsc{Oracle} representations is generally close across tasks, with only modest differences in AUPRC. The two are effectively identical for ICU Admission and remain similar for Acute MI, Lupus, Hypertension, Celiac, Pancreatic Cancer, and 30-day Readmission. Some larger differences are observed for Hyperlipidemia, Long LOS, and Sepsis Prediction, where the oracle representation performs differently from the rollout representation. Overall, the AUPRC evaluation complements the AUROC results by showing how predictive performance varies under substantial class imbalance while preserving the same task-dependent pattern across modeling approaches.

\subsection{Model Calibration}

\begin{figure}
    \centering
    \includegraphics[width=\linewidth]{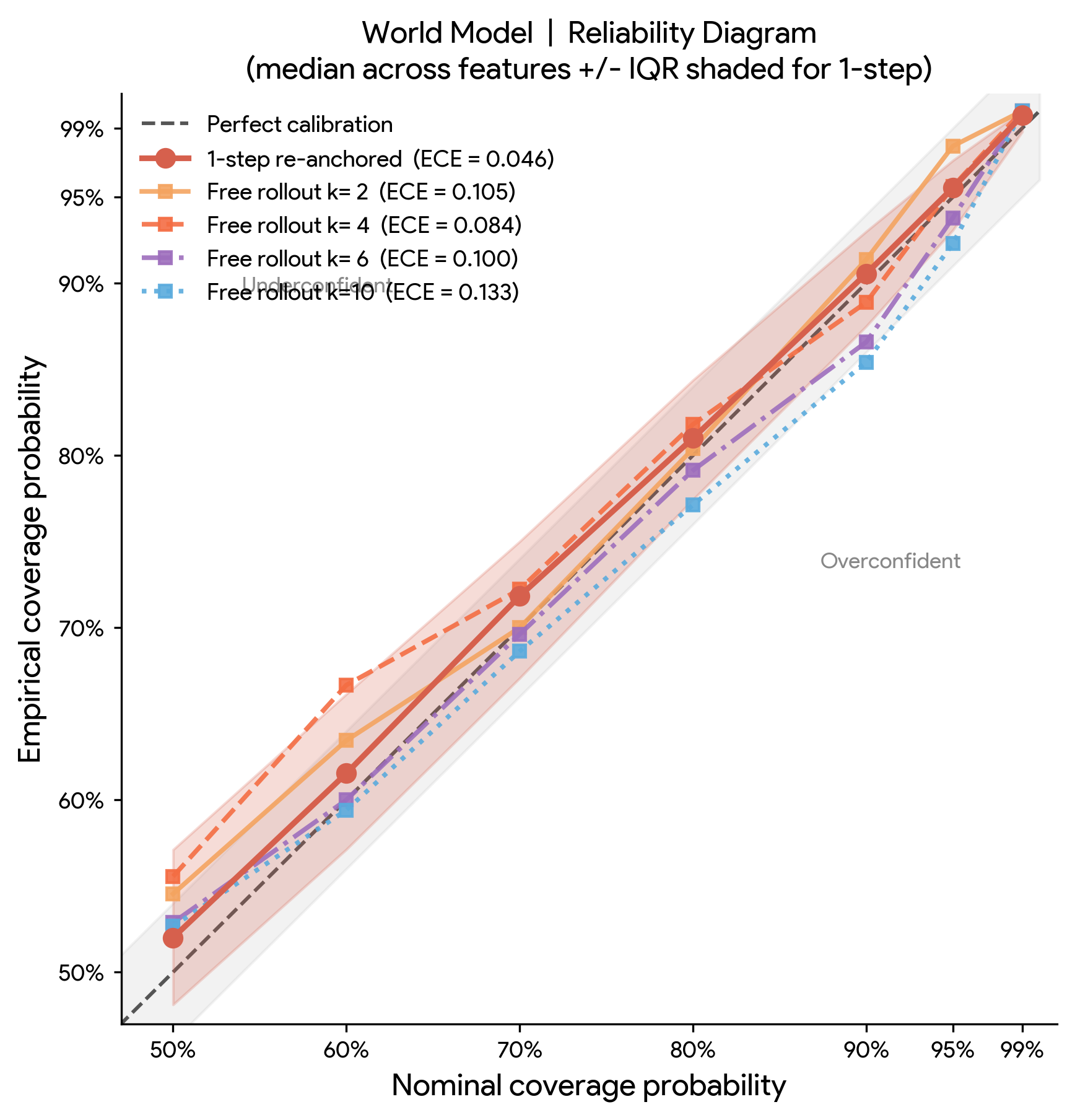}
    \caption{\small \textbf{Reliability of EHR-WM forecasts across rollout horizons.}
Reliability diagram comparing nominal prediction-interval coverage with empirical coverage for one-step re-anchored predictions and free-running rollouts at increasing horizons. The dashed diagonal denotes perfect calibration, while expected calibration error (ECE) summarizes deviation between nominal and empirical coverage.}
    \label{fig:placeholder}
\end{figure}

\begin{figure}
    \centering
    \includegraphics[width=\linewidth]{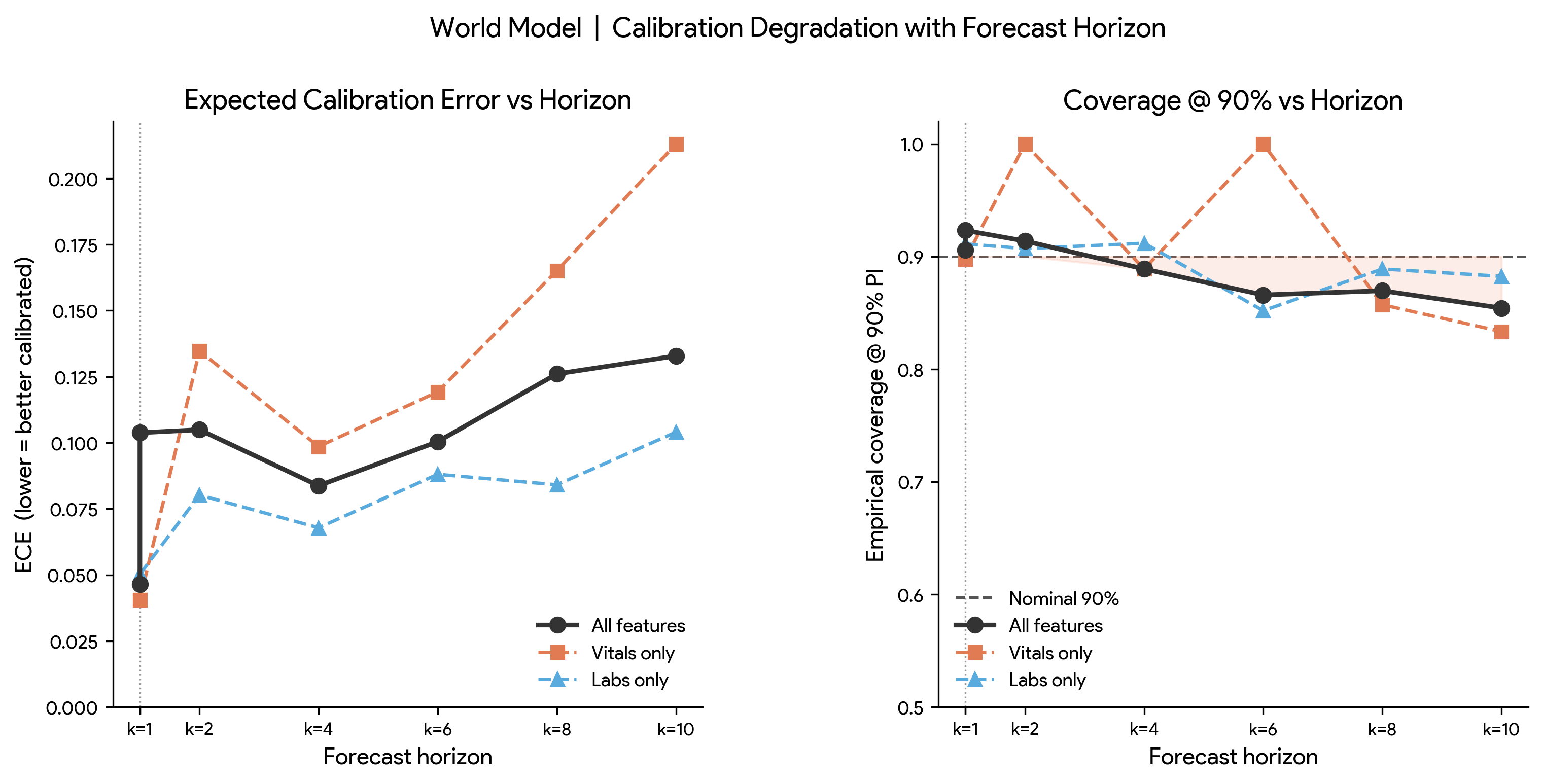}
    \caption{\small \textbf{Calibration degradation with increasing forecast horizon.}
Expected calibration error (left) and empirical coverage of the nominal 90\% prediction interval (right) are reported across rollout horizons for all continuous features, vital signs, and laboratory measurements. Increasing rollout depth generally produces larger calibration error and modest deviations from nominal coverage, characterizing uncertainty accumulation during multi-step forecasting.}
    \label{fig:placeholder}
\end{figure}

We further evaluate whether uncertainty associated with EHR-WM rollouts remains calibrated as predictions are propagated forward in time. The first figure presents a reliability analysis comparing nominal coverage probabilities with the empirical coverage achieved by the model. Perfect calibration corresponds to the diagonal, where an interval constructed to provide a given level of coverage contains the observed value at approximately that same frequency. We report a one-step re-anchored setting as a short-horizon reference and compare it with free-running rollouts at increasing horizons \(k\). This provides a direct view of whether predictive uncertainty remains representative of observed errors as the model transitions from local prediction to autonomous multi-step rollout. 

Across rollout horizons, the reliability curves remain broadly aligned with the nominal coverage levels, while deviations become more apparent as the model is propagated further from the observed trajectory. The expected calibration error (ECE) summarizes these deviations across coverage levels, providing a single measure of how closely empirical uncertainty agrees with the requested nominal coverage. The one-step re-anchored prediction provides the strongest calibration reference, while the free-rollout curves characterize the additional uncertainty introduced by repeated application of the learned transition dynamics.

The second figure examines this behavior explicitly as a function of forecast horizon. We report ECE across rollout lengths \(k=1,\ldots,10\), together with empirical coverage at the nominal \(90\%\) level. Results are shown for all continuous features jointly and separately for vital signs and laboratory measurements. This decomposition allows calibration degradation to be examined both globally and across clinically distinct measurement groups, since different modalities can exhibit different levels of observation noise, sparsity, and temporal predictability. 

Together, the two views characterize complementary aspects of uncertainty under rollout. The reliability diagram evaluates the full relationship between nominal and empirical coverage at selected rollout depths, whereas the horizon analysis summarizes how calibration changes as predictions are propagated progressively further into the future. This provides an additional evaluation of the world model beyond point-estimate metrics such as \(R^2\) and MAE, assessing not only whether future clinical values can be forecast, but also whether the uncertainty associated with those forecasts remains meaningful during autonomous rollout.

\clearpage
\newpage
\subsection{Odds Ratios and Hazard Ratios by Demographic Groups}

\begin{figure*}[h!]
    \centering
    \includegraphics[width=0.8\linewidth]{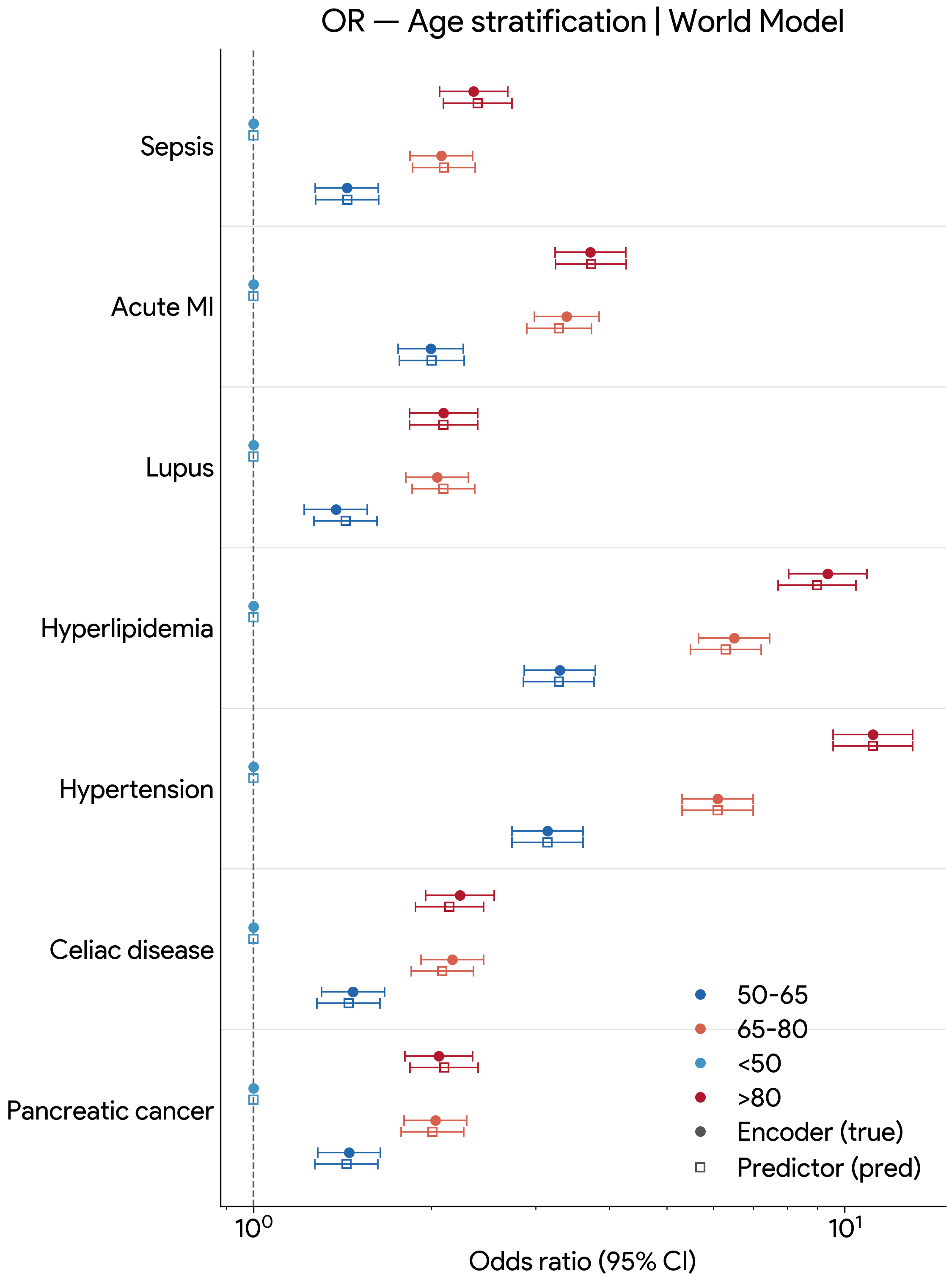}
    \caption{\small \textbf{Age-stratified odds ratios from generated trajectories.}
Odds ratios (95\% CI) computed from predictor-generated trajectories closely match those from encoder-derived representations across age groups, preserving relative risk structure and clinically consistent trends.}
    \label{fig:placeholder}
\end{figure*}

\begin{figure*}
    \centering
    \includegraphics[width=0.8\linewidth]{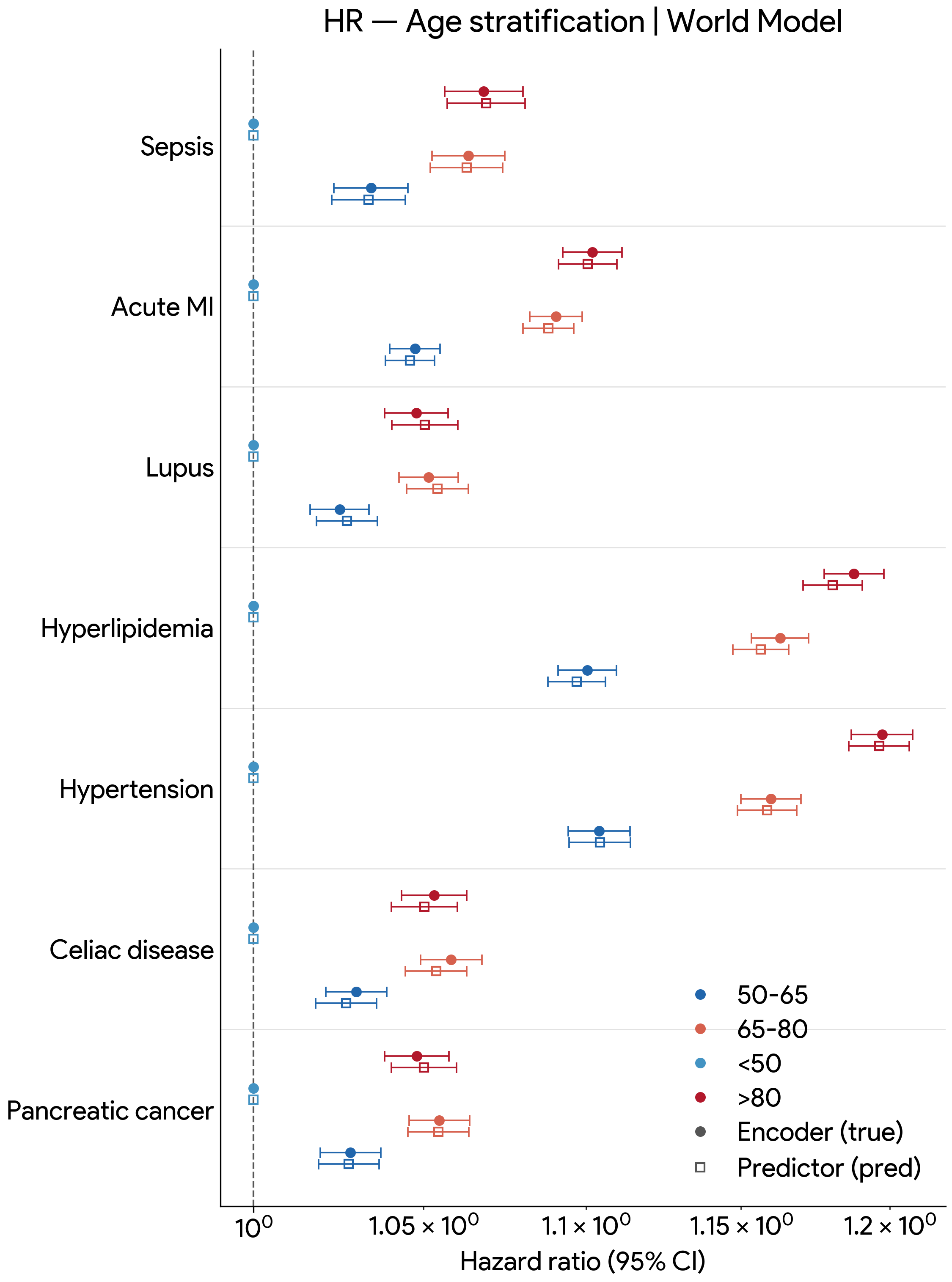}
    \caption{\textbf{Age-stratified hazard ratios from generated trajectories.}
Hazard ratios (95\% CI) computed from predictor-generated trajectories closely match those from encoder-derived representations across age groups, preserving relative risk structure and clinically consistent trends.}
    \label{fig:placeholder}
\end{figure*}

\begin{figure*}
    \centering
    \includegraphics[width=0.8\linewidth]{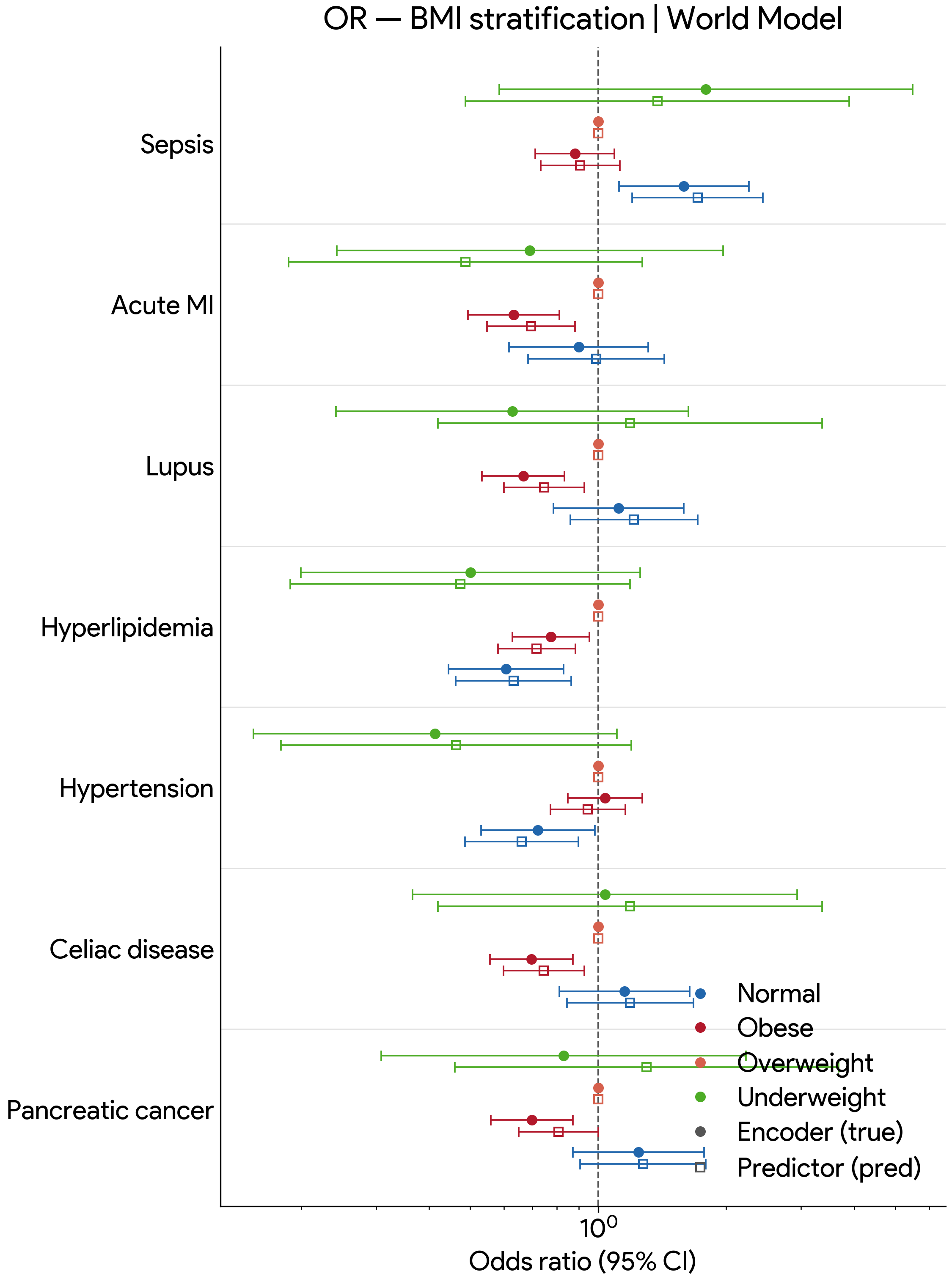}
    \caption{\textbf{BMI-stratified odds ratios from generated trajectories.}
Odds ratios (95\% CI) computed from predictor-generated trajectories closely match those from encoder-derived representations across BMI groups, preserving relative risk structure and clinically consistent trends.}
    \label{fig:placeholder}
\end{figure*}

\begin{figure*}
    \centering
    \includegraphics[width=0.8\linewidth]{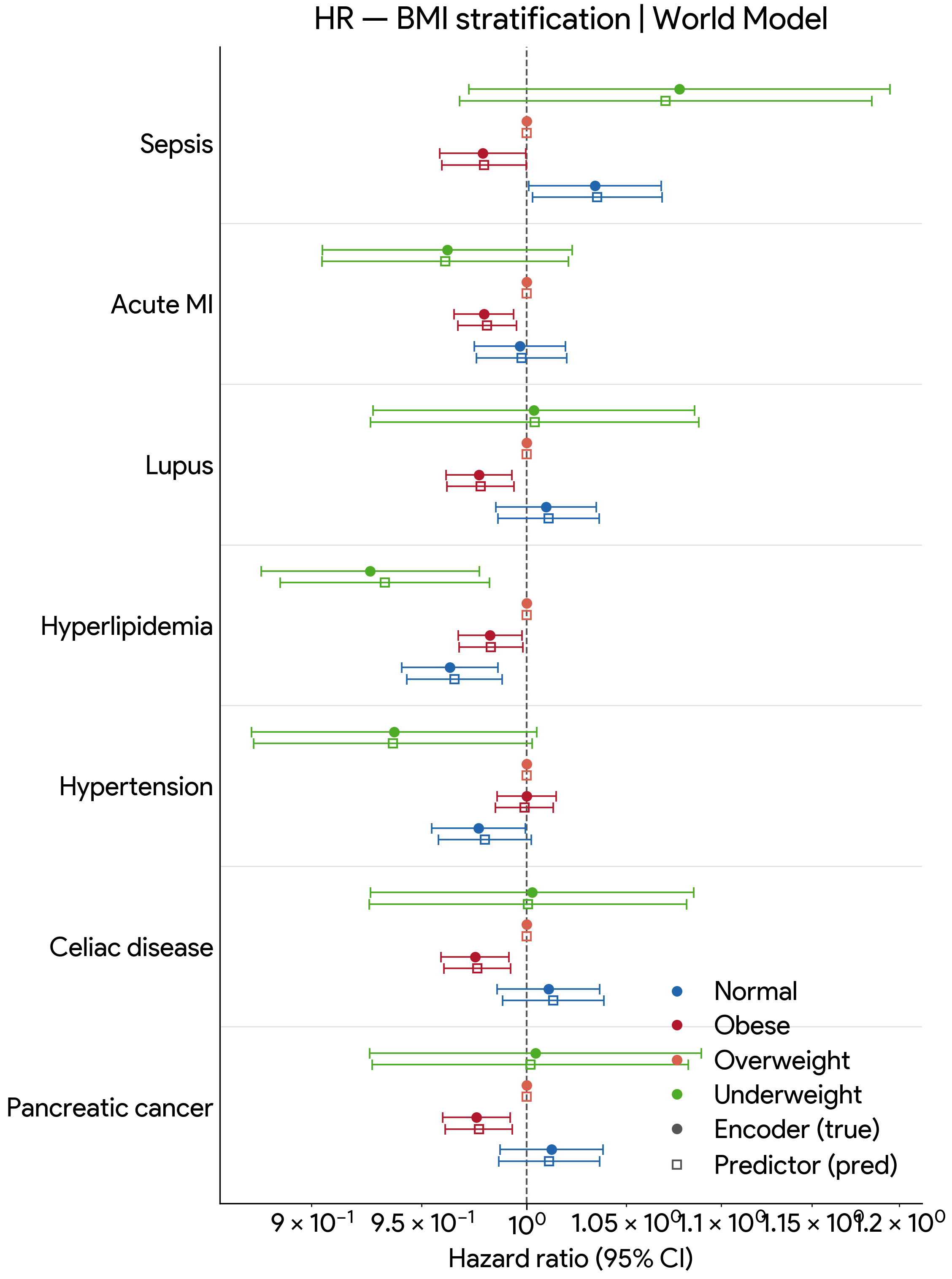}
    \caption{\textbf{BMI-stratified hazard ratios from generated trajectories.}
Hazard ratios (95\% CI) computed from predictor-generated trajectories closely match those from encoder-derived representations across BMI groups, preserving relative risk structure and clinically consistent trends.}
    \label{fig:placeholder}
\end{figure*}

\begin{figure*}
    \centering
    \includegraphics[width=0.8\linewidth]{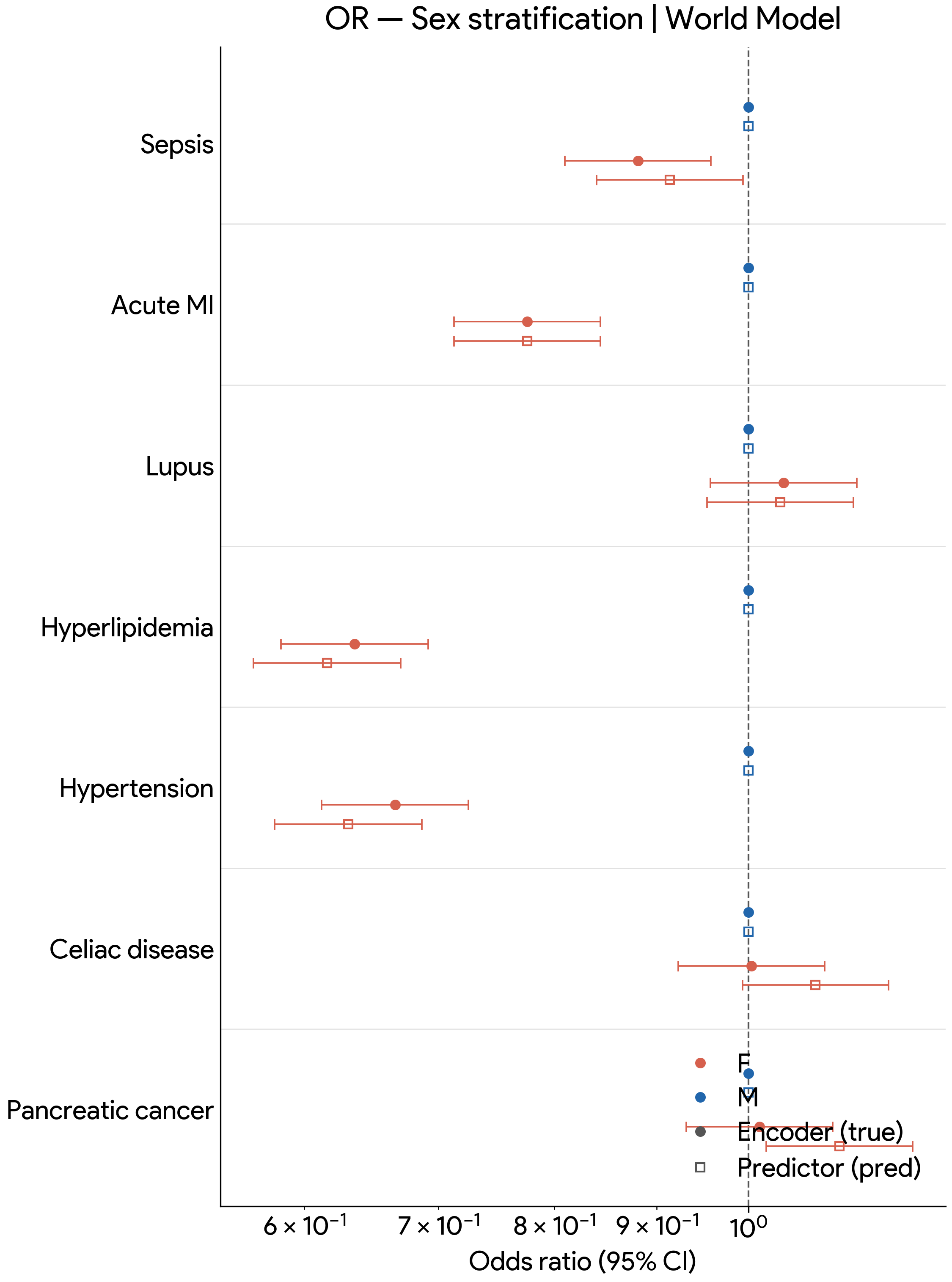}
    \caption{\textbf{Sex-stratified odds ratios from generated trajectories.}
Odds ratios (95\% CI) computed from predictor-generated trajectories closely match those from encoder-derived representations across sex groups, preserving relative risk structure and clinically consistent trends.}
    \label{fig:placeholder}
\end{figure*}

\begin{figure*}
    \centering
    \includegraphics[width=0.8\linewidth]{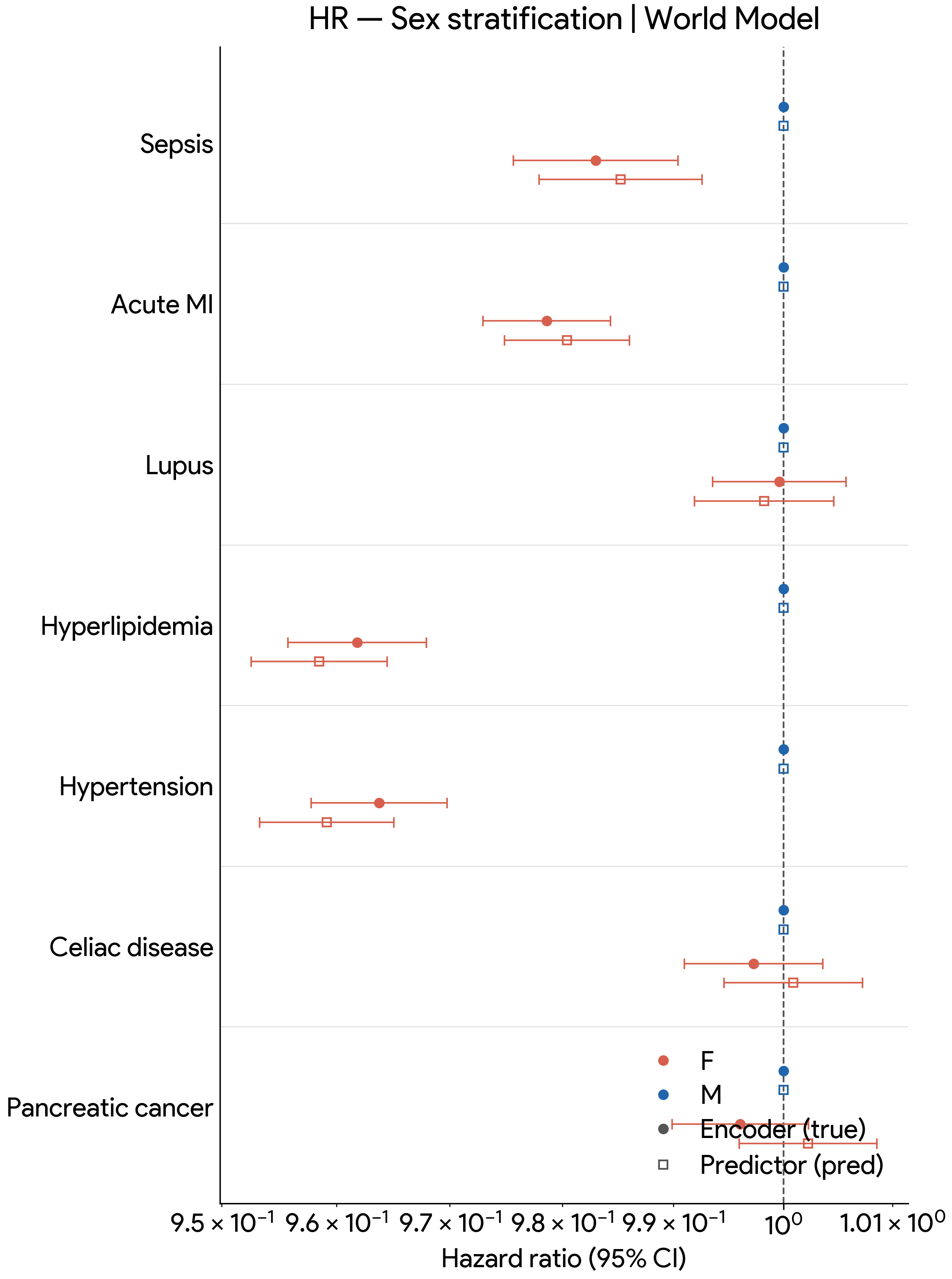}
    \caption{\textbf{Sex-stratified hazard ratios from generated trajectories.}
Hazard ratios (95\% CI) computed from predictor-generated trajectories closely match those from encoder-derived representations across sex groups, preserving relative risk structure and clinically consistent trends.}
    \label{fig:placeholder}
\end{figure*}

\newpage
\subsection{Temporal Probing}
\label{tp}

\begin{figure*}[h!]
    \centering
    \includegraphics[width=\linewidth]{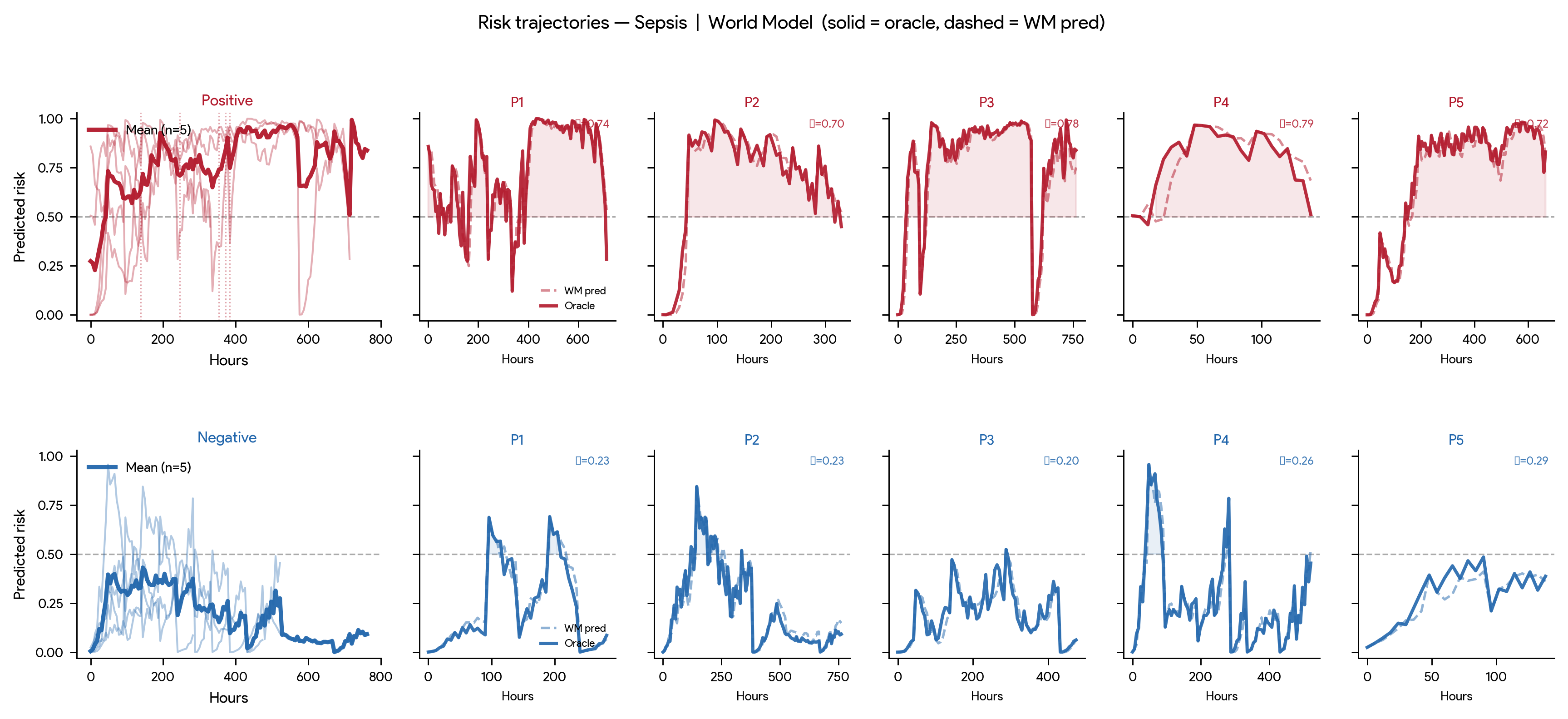}
    \caption{\small \textbf{Trajectory-level risk evolution under world model rollouts.} Predicted risk trajectories for sepsis comparing encoder-derived (``oracle'', solid) and world model rollout (``WM pred'', dashed) representations. Top row shows positive cases, bottom row negative cases, with cohort means (left) and representative patients (P1–P5). The model closely tracks temporal risk dynamics, capturing both sharp transitions and longer-term trends, while maintaining consistency with oracle trajectories across diverse clinical patterns.}
    \label{fig:placeholder}
\end{figure*}

\begin{figure*}[h!]
    \centering
    \includegraphics[width=\linewidth]{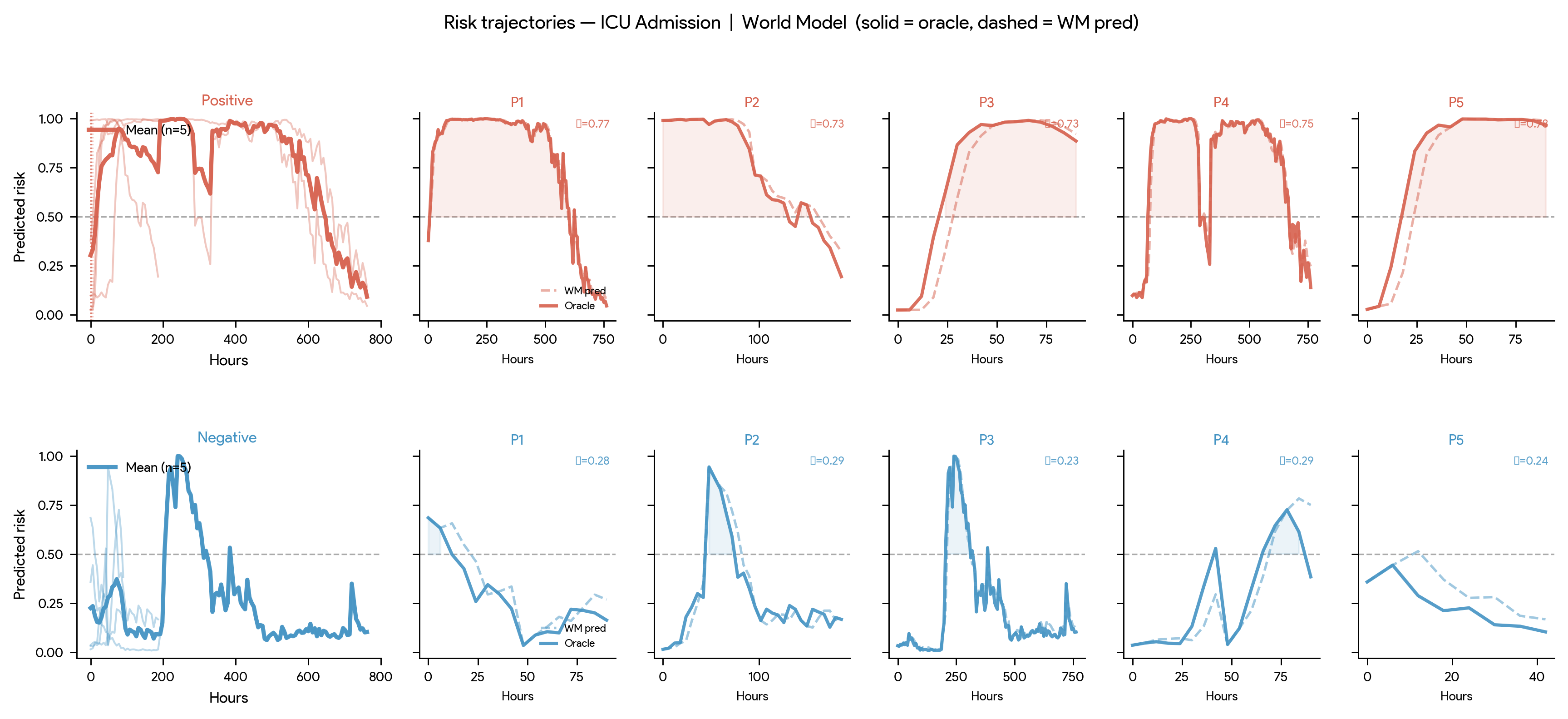}
    \caption{\small\textbf{Trajectory-level risk evolution under world model rollouts.} Predicted risk trajectories for ICU Admission comparing encoder-derived (``oracle'', solid) and world model rollout (``WM pred'', dashed) representations. Top row shows positive cases, bottom row negative cases, with cohort means (left) and representative patients (P1–P5). The model closely tracks temporal risk dynamics, capturing both sharp transitions and longer-term trends, while maintaining consistency with oracle trajectories across diverse clinical patterns.}
    \label{fig:placeholder}
\end{figure*}

\newpage
\subsection{Forecasting MAE and $R^2$}
\label{fore1}

\begin{figure*}[h!]
    \centering
    \includegraphics[width=0.8\linewidth]{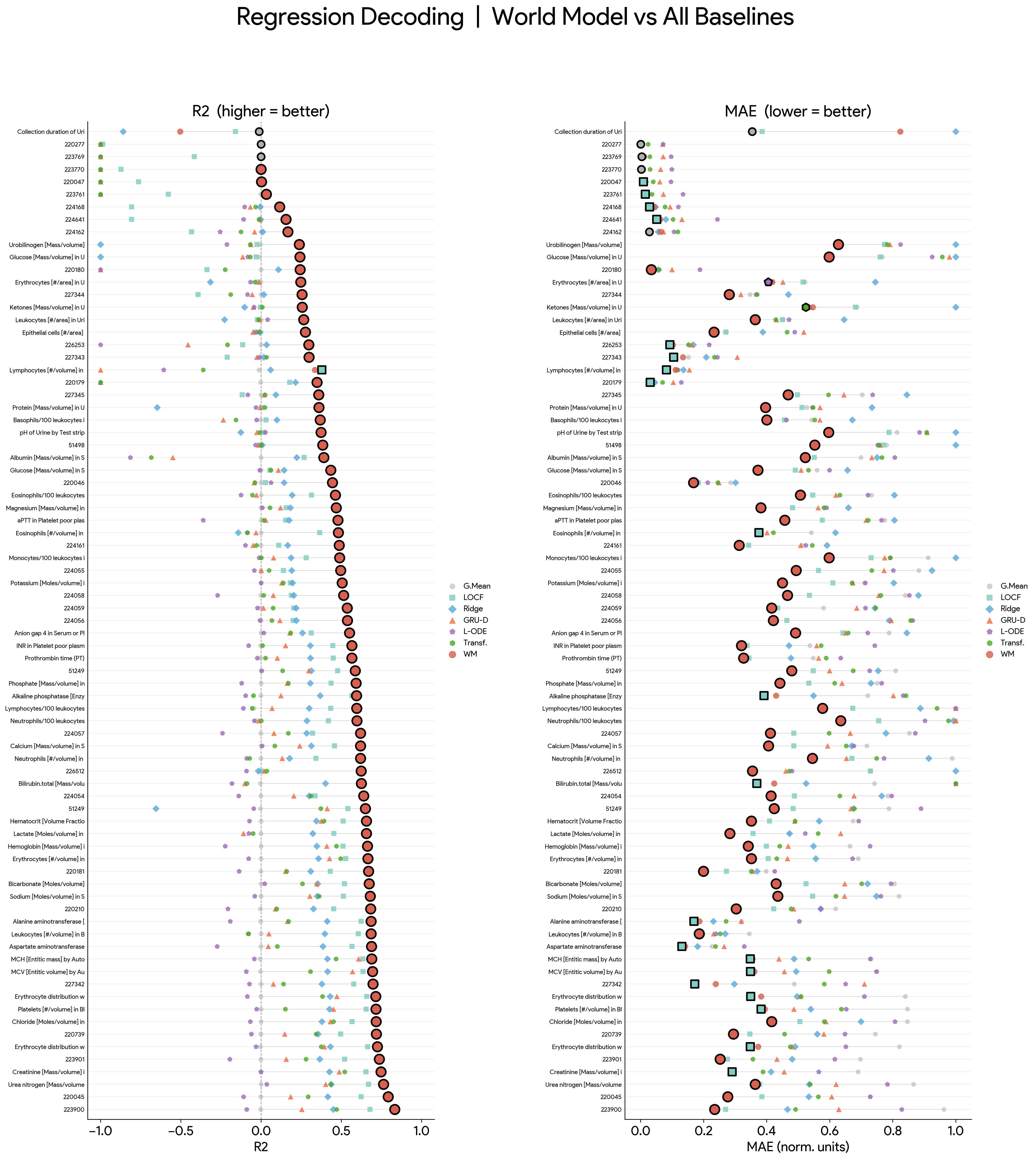}
    \caption{\small \textbf{Regression decoding performance across clinical variables.} Per-feature evaluation of forecasting accuracy comparing the world model (WM) against global mean and last observation carried forward (LOCF) baselines. Left: coefficient of determination ($R^2$); right: mean absolute error (MAE). Each point corresponds to a clinical variable. The world model consistently achieves higher $R^2$ and lower MAE across most features, indicating improved fidelity in capturing temporal dynamics and reconstructing future observations relative to static or persistence-based baselines.}
    \label{fig:placeholder}
\end{figure*}

\newpage
\subsection{Time Window Regression}
\label{fore2}

\begin{figure*}[h!]
    \centering
    \includegraphics[width=\linewidth]{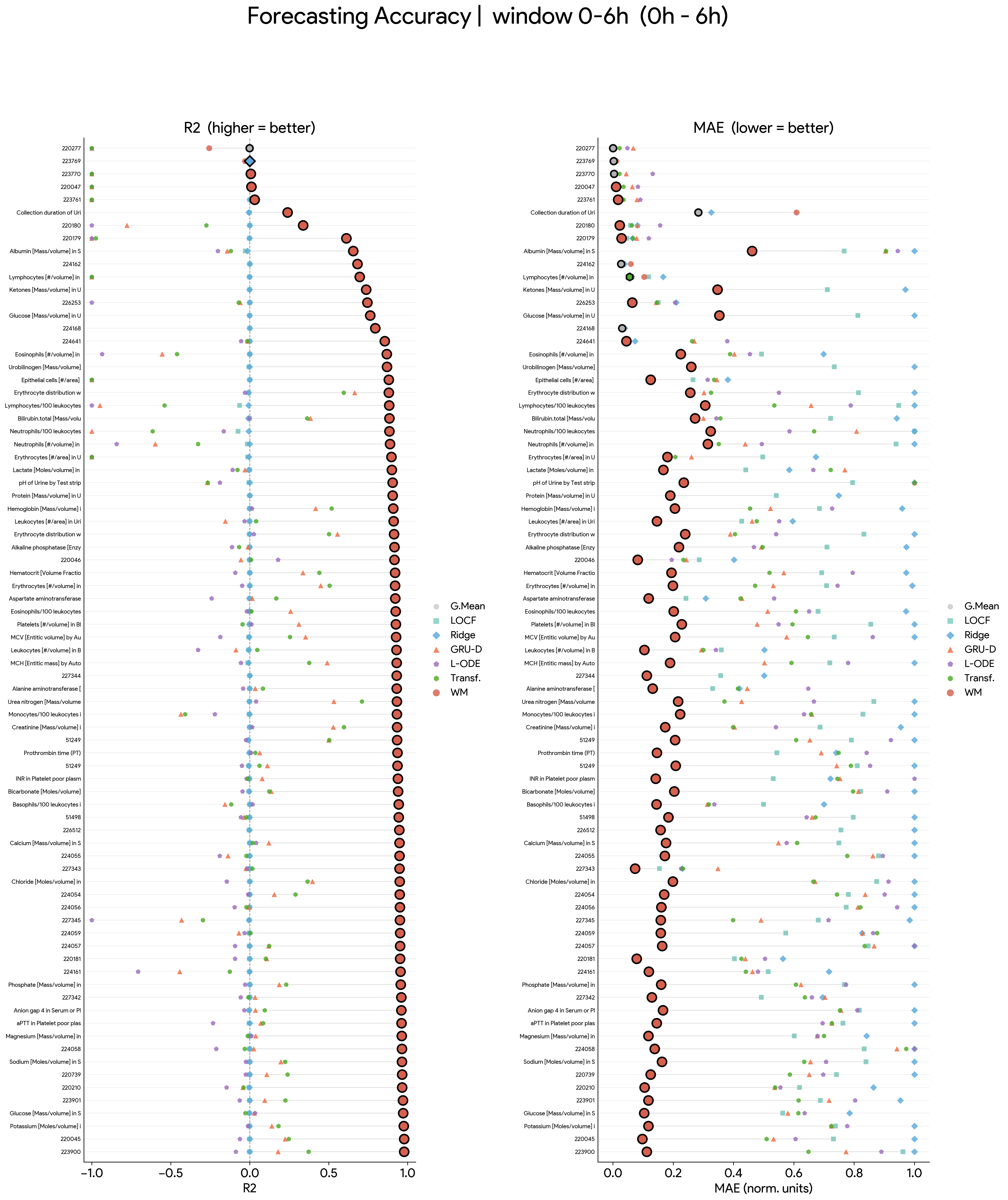}
    \caption{\small \textbf{Regression decoding performance at long-horizon rollout (0–6h).} Per-feature evaluation of forecasting accuracy for the 0–6 hour prediction window, with features sorted by world model ($\text{WM}$) $R^2$ in descending order. Left: coefficient of determination ($R^2$); right: mean absolute error (MAE).}
    \label{fig:placeholder}
\end{figure*}

\begin{figure*}
    \centering
    \includegraphics[width=\linewidth]{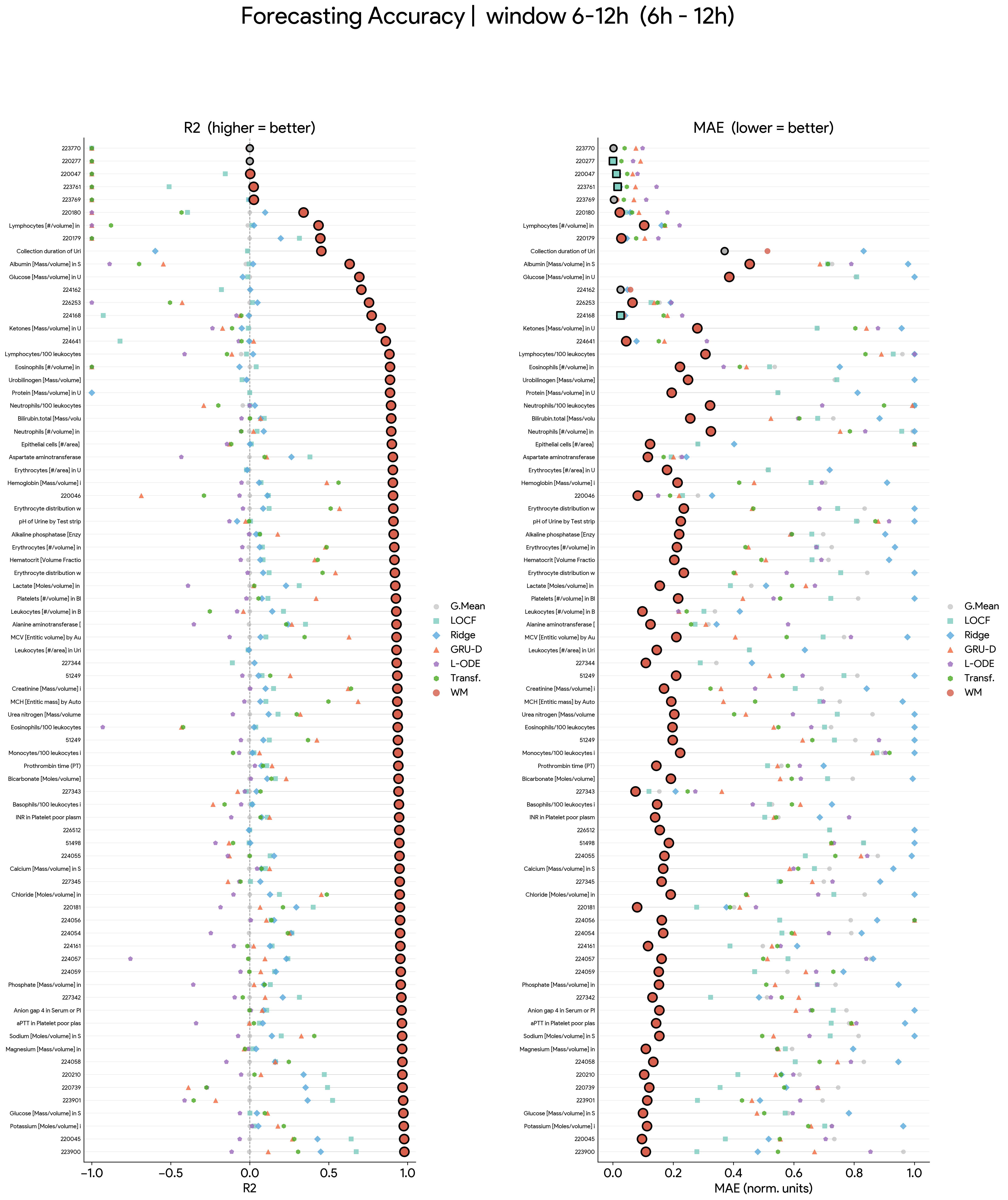}
    \caption{\small \textbf{Regression decoding performance at long-horizon rollout (6–12h).} Per-feature evaluation of forecasting accuracy for the 6–12 hour prediction window, with features sorted by world model ($\text{WM}$) $R^2$ in descending order. Left: coefficient of determination ($R^2$); right: mean absolute error (MAE).}
    \label{fig:placeholder}
\end{figure*}

\begin{figure*}
    \centering
    \includegraphics[width=\linewidth]{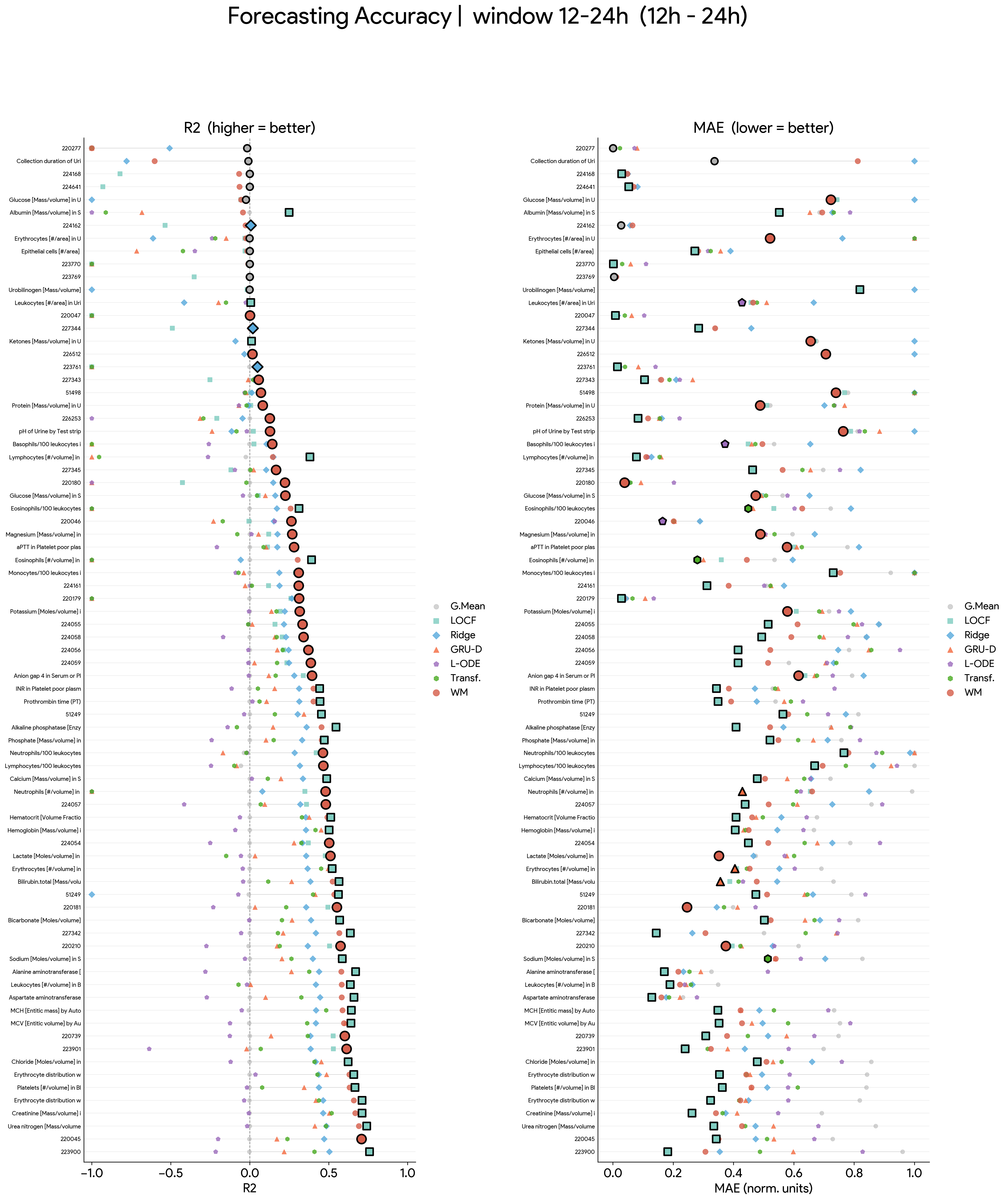}
    \caption{\small \textbf{Regression decoding performance at long-horizon rollout (12–24h).} Per-feature evaluation of forecasting accuracy for the 12–24 hour prediction window, with features sorted by world model ($\text{WM}$) $R^2$ in descending order. Left: coefficient of determination ($R^2$); right: mean absolute error (MAE).}
    \label{fig:placeholder}
\end{figure*}

\begin{figure*}
    \centering
    \includegraphics[width=\linewidth]{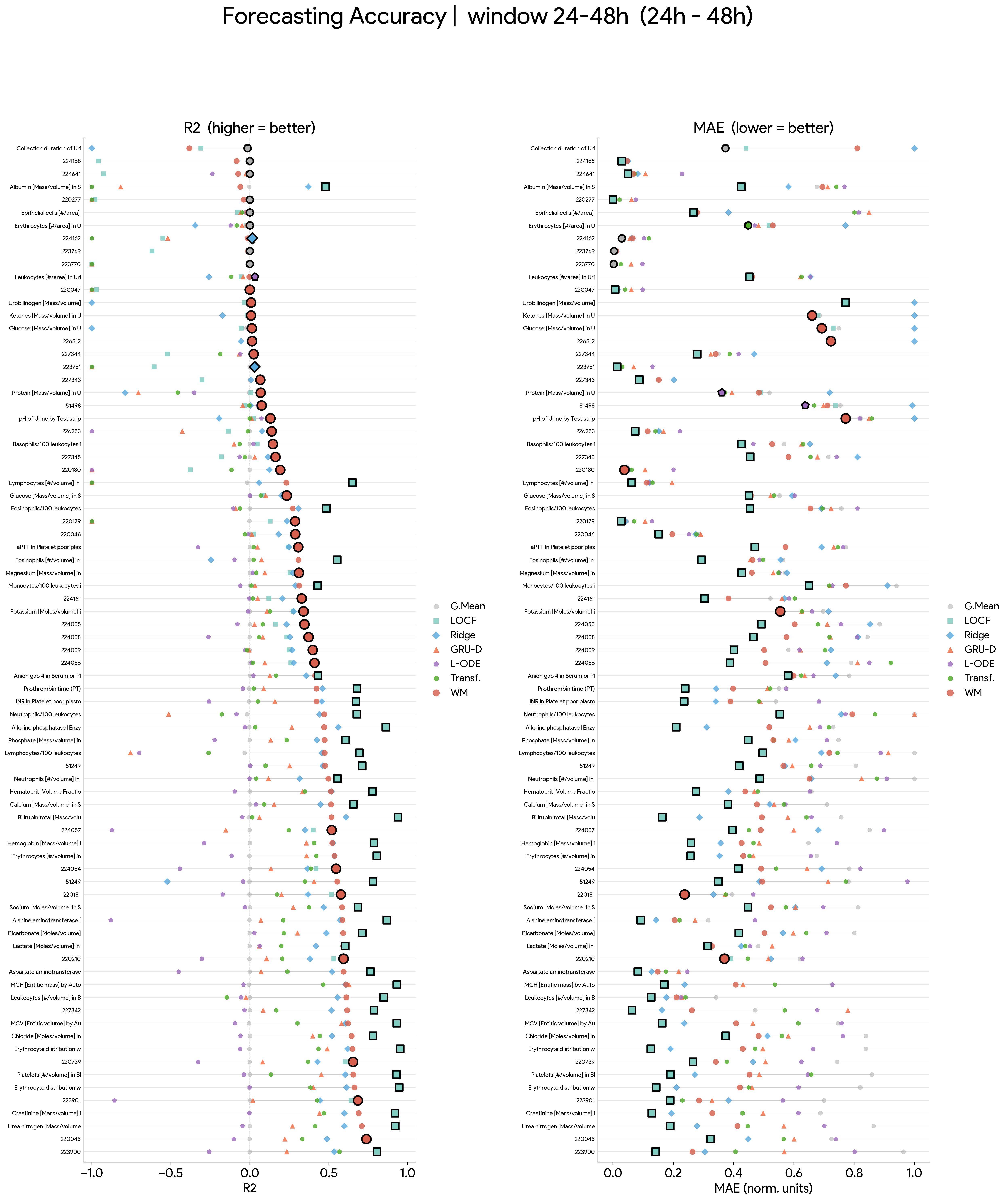}
    \caption{\small \textbf{Regression decoding performance at long-horizon rollout (24–48h).} Per-feature evaluation of forecasting accuracy for the 24–48 hour prediction window, with features sorted by world model ($\text{WM}$) $R^2$ in descending order. Left: coefficient of determination ($R^2$); right: mean absolute error (MAE).}
    \label{fig:placeholder}
\end{figure*}

\end{document}